%% file: main.tex
\documentclass[journal,twoside]{IEEEtran}

\usepackage{cite}

\ifCLASSINFOpdf%
  \usepackage[pdftex]{graphicx}
  \graphicspath{{imgs/}}
\else
\fi

\usepackage{microtype}

\usepackage{amsmath}
\usepackage{amssymb}
\usepackage{mathtools}

\usepackage{bm}

\DeclareMathOperator*{\argmin}{arg\,min}
\DeclareMathOperator*{\argmax}{arg\,max}

 \usepackage[caption=false,font=footnotesize]{subfig}
\usepackage[hyphens]{url} %% Embedding URL's in document.
\usepackage{hyperref}
\hypersetup{
    colorlinks,
    linkcolor  = {red!80!black},
    citecolor  = {blue!80!black},
    urlcolor   = {blue!20!black}
}

\usepackage{booktabs}
\usepackage{multirow}
\usepackage{tabularx}
\usepackage{threeparttable}

\newcommand{\supcite}[1]{\textsuperscript{\cite{#1}}}

\usepackage{xcolor}

\let\oldleft\left
\let\oldright\right
\renewcommand{\left}{\mathopen{}\mathclose\bgroup\oldleft}
\renewcommand{\right}{\aftergroup\egroup\oldright}

\newcommand{\rbracket}[1]{\left(#1\right)}
\newcommand{\sbracket}[1]{\left[#1\right]}
\newcommand{\cbracket}[1]{\left\{#1\right\}}

\newcommand{\tablehead}[1]{\multicolumn{1}{c}{\bfseries#1}}
\newcommand{\tableheadwithvline}[1]{\multicolumn{1}{c|}{\bfseries#1}}

\usepackage{tikz}
\usetikzlibrary{arrows.meta,calc,decorations.pathreplacing}

\definecolor{C0}{HTML}{1F77B4}
\definecolor{C1}{HTML}{FF7F0E}
\definecolor{C2}{HTML}{2CA02C}

\begin{document}

\title{A Survey on Neural Network Interpretability}

\author{Yu~Zhang,
Peter~Ti\v{n}o,
Ale\v{s}~Leonardis,
and~Ke~Tang%
\thanks{This work was supported in part by the Guangdong Provincial Key Laboratory under Grant 2020B121201001;
in part by the Program for Guangdong Introducing Innovative and Entrepreneurial Teams under Grant 2017ZT07X386;
in part by the Stable Support Plan Program of Shenzhen Natural Science Fund under Grant 20200925154942002;
in part by the Science and Technology Commission of Shanghai Municipality under Grant 19511120602;
in part by the National Leading Youth Talent Support Program of China;
and in part by the MOE University Scientific-Technological Innovation Plan Program.
Peter Tino was supported by the European Commission Horizon 2020 Innovative Training Network SUNDIAL (SUrvey Network for Deep Imaging Analysis and Learning), Project ID: 721463.
We also acknowledge MoD/Dstl and EPSRC for providing the grant to support the UK academics involvement in a Department of Defense funded MURI project through EPSRC grant EP/N019415/1.}%
\thanks{Y.~Zhang and K.~Tang are with the Guangdong Key Laboratory of Brain-Inspired Intelligent Computation, Department of Computer Science and Engineering,
Southern University of Science and Technology, Shenzhen 518055, P.R.China, and also with the Research Institute of Trust-worthy Autonomous Systems,
Southern University of Science and Technology, Shenzhen 518055, P.R.China (e-mail: zhangy3@mail.sustech.edu.cn, tangk3@sustech.edu.cn).}% <-this % stops a space
\thanks{Y.~Zhang, P.~Ti\v{n}o and A.~Leonardis are with the School of Computer Science,
University of Birmingham, Edgbaston, Birmingham B15 2TT, UK (e-mail: \{p.tino, a.leonardis\}@cs.bham.ac.uk).}%
\thanks{\textit{Manuscript accepted July 09, 2021, IEEE-TETCI.
© 2021 IEEE. Personal use of this material is permitted.
Permission from IEEE must be obtained for all other uses, in any current or future media, including reprinting/republishing this material for advertising or promotional purposes,
creating new collective works, for resale or redistribution to servers or lists, or reuse of any copyrighted component of this work in other works.}}%
}

%% The paper headers
\markboth{IEEE Transactions on XXXX,~Vol.~X, No.~X, MM~YYYY}%
{Zhang \MakeLowercase{\textit{et al.}}: A Survey on Neural Network Interpretability}
% The only time the second header will appear is for the odd numbered pages
% after the title page when using the twoside option.
% 
% *** Note that you probably will NOT want to include the author's ***
% *** name in the headers of peer review papers.                   ***
% You can use \ifCLASSOPTIONpeerreview for conditional compilation here if
% you desire.

% If you want to put a publisher's ID mark on the page you can do it like
% this:
%\IEEEpubid{0000--0000/00\$00.00~\copyright~2015 IEEE}
% Remember, if you use this you must call \IEEEpubidadjcol in the second
% column for its text to clear the IEEEpubid mark.

% use for special paper notices
%\IEEEspecialpapernotice{(Invited Paper)}

% make the title area
\maketitle

% As a general rule, do not put math, special symbols or citations
% in the abstract or keywords.
\begin{abstract}
Along with the great success of deep neural networks, there is also growing concern about their black-box nature.
The interpretability issue affects people's trust on deep learning systems.
It is also related to many ethical problems, e.g., algorithmic discrimination.
Moreover, interpretability is a desired property for deep networks to become powerful tools in other research fields, e.g., drug discovery and genomics.
In this survey, we conduct a comprehensive review of the neural network interpretability research.
We first clarify the definition of interpretability as it has been used in many different contexts.
Then we elaborate on the importance of interpretability and propose a novel taxonomy organized along three dimensions: type of engagement (passive vs.\ active interpretation approaches), the type of explanation, and the focus (from local to global interpretability).
This taxonomy provides a meaningful 3D view of distribution of papers from the relevant literature as two of the dimensions are not simply categorical but allow ordinal subcategories.
Finally, we summarize the existing interpretability evaluation methods and suggest possible research directions inspired by our new taxonomy.
\end{abstract}

% Note that keywords are not normally used for peerreview papers.

% For peer review papers, you can put extra information on the cover
% page as needed:
% \ifCLASSOPTIONpeerreview
% \begin{center} \bfseries EDICS Category: 3-BBND \end{center}
% \fi
%
% For peerreview papers, this IEEEtran command inserts a page break and
% creates the second title. It will be ignored for other modes.
\IEEEpeerreviewmaketitle

\section{Introduction}\label{sec:introduction}

\IEEEPARstart{O}{ver} the last few years, deep neural networks (DNNs) have achieved tremendous success~\cite{lecun2015dl} in computer vision~\cite{2012icwdc, 2016drlfi}, speech recognition~\cite{hinton2012dnnfa}, natural language processing~\cite{2014stslw} and other fields~\cite{2017mtgog},
while the latest applications can be found in these surveys~\cite{dong2021survey,dixit2021deep,bouwmans2019deep}.
They have not only beaten many previous machine learning techniques (e.g., decision trees, support vector machines), but also achieved the state-of-the-art performance on certain real-world tasks\cite{wu2016gnmts,hinton2012dnnfa}.
Products powered by DNNs are now used by billions of people\footnote{\url{https://www.acm.org/media-center/2019/march/turing-award-2018}}, e.g., in facial and voice recognition.
DNNs have also become powerful tools for many scientific fields, such as medicine~\cite{wainberg2018dlib}, bioinformatics~\cite{xiong2015thscr,2017tptis} and astronomy\cite{parks2018dloqs}, which usually involve massive data volumes.

However, deep learning still has some significant disadvantages.
As a really complicated model with millions of free parameters (e.g., AlexNet~\cite{2012icwdc}, 62 million), DNNs are often found to exhibit unexpected behaviours.
For instance, even though a network could get the state-of-the-art performance and seemed to generalize well on the object recognition task, Szegedy et al.~\cite{szegedy2013iponn} found a way that could arbitrarily change the network's prediction by applying a certain imperceptible change to the input image.
This kind of modified input is called ``adversarial example''.
Nguyen et al.~\cite{nguyen2015dnnae} showed another way to produce completely unrecognizable images (e.g., look like white noise), which are, however, recognized as certain objects by DNNs with 99.99\% confidence.
These observations suggest that even though DNNs can achieve superior performance on many tasks, their underlying mechanisms may be very different from those of humans' and have not yet been well-understood.

\subsection{An (Extended) Definition of Interpretability}\label{sec:interpretability}

To open the black-boxes of deep networks, many researchers started to focus on the model interpretability.
Although this theme has been explored in various papers, no clear consensus on the definition of interpretability has been reached.
Most previous works skimmed over the clarification issue and left it as ``you will know it when you see it''.
If we take a closer look, the suggested definitions and motivations for interpretability are often different or even discordant~\cite{lipton2016tmomi}.

One previous definition of interpretability is \textit{the ability to provide explanations in understandable terms to a human}~\cite{doshi2017tarso}, while the term \textit{explanation} itself is still elusive.
After reviewing previous literature, we make further clarifications of ``explanations'' and ``understandable terms'' on the basis of~\cite{doshi2017tarso}.

\begin{quote}
    \textbf{Interpretability} is the ability to provide \textit{explanations}\textsuperscript{1} in \textit{understandable terms}\textsuperscript{2} to a human.
\end{quote}
where
\begin{enumerate}
    \item \textbf{Explanations}, ideally, should be \textit{logical decision rules} (if-then rules) or can be transformed to logical rules.
          However, people usually do not require explanations to be explicitly in a rule form (but only some key elements which can be used to construct explanations).
    \item \textbf{Understandable terms} should be from the \textit{domain knowledge} related to the task (or common knowledge according to the task).
\end{enumerate}

Our definition enables new perspectives on the interpretability research:
{\bf (1)} We highlight \textit{the form of explanations} rather than particular \textit{explanators}.
After all, explanations are expressed in a certain ``language'', be it natural language, logic rules or something else.
{Recently, a strong preference has been expressed for the language of explanations to be as close as possible to logic~\cite{pedreschi2019meobb}.}
In practice, people do not always require a full ``sentence'', which allows various kinds of explanations (rules, saliency masks etc.).
This is an important angle to categorize the approaches in the existing literature.
{\bf (2)} Domain knowledge is the basic unit in the construction of explanations.
As deep learning has shown its ability to process data in the raw form, it becomes harder for people to interpret the model with its original input representation.
With more domain knowledge, we can get more understandable representations that can be evaluated by domain experts.
Table~\ref{tab:terms} lists several commonly used representations in different tasks.

\begin{table}[!t]
    \centering
    \begin{threeparttable}
        \scriptsize
        \renewcommand{\arraystretch}{1.3}
        \caption{Some interpretable ``terms'' used in practice.}\label{tab:terms}
        \begin{tabular}{lll}
            \toprule
            \tablehead{Field}                & \tablehead{Raw input}            & \tablehead{Understandable terms}         \\
            \midrule
            \multirow{2}{*}{Computer vision} & \multirow{2}{*}{Images (pixels)} & Super pixels (image patches)\tnote{a}    \\
                                             &                                  & Visual concepts\tnote{b}                 \\
            NLP                              & Word embeddings                  & Words                                    \\
            Bioinformatics                   & Sequences                        & Motifs (position weight matrix)\tnote{c} \\
            \bottomrule
        \end{tabular}
        \begin{tablenotes}
            \item [a] image patches are usually used in attribution methods~\cite{ribeiro2016wsity}.
            \item [b] colours, materials, textures, parts, objects and scenes~\cite{bau2017ndqio}.
            \item [c] proposed by~\cite{1982uotpa} and became an essential tool for computational motif discovery.
        \end{tablenotes}
    \end{threeparttable}
\end{table}

{We note that some studies distinguish between \textit{interpretability} and \textit{explainability} (or \textit{understandability}, \textit{comprehensibility}, \textit{transparency}, \textit{human-simulatability} etc.~\cite{lipton2016tmomi,arrieta2020eaixc}).
In this paper we do not emphasize the subtle differences among those terms.
As defined above, we see explanations as the core of interpretability and use interpretability, explainability and understandability interchangeably.
Specifically, we focus on the interpretability of (deep) \textit{neural networks} (rarely recurrent neural networks), which aims to provide explanations of their inner workings and input-output mappings.}
There are also some interpretability studies about the Generative Adversarial Networks (GANs).
However, as a kind of generative models, it is slightly different from common neural networks used as discriminative models.
For this topic, we would like to refer readers to the latest work~\cite{bau2019gandissect,yang2021semantic,voynov2020unsupervised,plumerault2020controlling,harkonen2020ganspace,kahng2018gan},
many of which share the similar ideas with the ``hidden semantics'' part of this paper (see Section~\ref{sec:taxonomy}), trying to interpret the meaning of hidden neurons or the latent space.

Under our definition, the source code of Linux operating system is interpretable although it might be overwhelming for a developer.
A deep decision tree or a high-dimensional linear model (on top of interpretable input representations) are also interpretable.
One may argue that they are not simulatable~\cite{lipton2016tmomi} (i.e.\ a human is able to simulate the model's processing from input to output in his/her mind in a short time).
We claim, however, they are still interpretable.

Besides above confined scope of interpretability (of a trained neural network), there is a much broader field of understanding the general neural network methodology, which cannot be covered by this paper.
For example, the empirical success of DNNs raises many unsolved questions to theoreticians~\cite{vidal2017mathematics}.
What are the merits (or inductive bias) of DNN architectures~\cite{bruna2013invariant,eldan2016power}?
What are the properties of DNNs' loss surface/critical points~\cite{nouiehed2018learning,yun2018global,choromanska2015loss,haeffele2017global}?
Why DNNs generalizes so well with just simple regularization~\cite{srivastava2014dropout,mianjy2018implicit,salehinejad2019ising}?
What about DNNs' robustness/stability~\cite{sengupta2018robust,zheng2016improving,haber2017stable,chang2018reversible,thekumparampil2018robustness,creswell2018denoising}?
There are also studies about how to generate adversarial examples~\cite{mopuri2017fast,mopuri2018nag} and detect adversarial inputs~\cite{zheng2018robust}.

\subsection{The Importance of Interpretability}

The need for interpretability has already been stressed by many papers~\cite{lipton2016tmomi,doshi2017tarso,guidotti2018asomf}, emphasizing cases where lack of interpretability may be harmful.
However, a clearly organized exposition of such argumentation is missing.
We summarize the arguments for the importance of interpretability into three groups.

\subsubsection{High Reliability Requirement}

Although deep networks have shown great performance on some relatively large test sets, the real world environment is still much more complex.
As some unexpected failures are inevitable, we need some means of making sure we are still in control.
Deep neural networks do not provide such an option.
In practice, they have often been observed to have unexpected performance drop in certain situations, not to mention the potential attacks from the adversarial examples~\cite{2016tlodl,2017uap}.

Interpretability is not always needed but it is important for some prediction systems that are required to be highly reliable because an error may cause catastrophic results (e.g., human lives, heavy financial loss).
Interpretability can make potential failures easier to detect (with the help of domain knowledge), avoiding severe consequences.
Moreover, it can help engineers pinpoint the root cause and provide a fix accordingly.
Interpretability does not make a model more reliable or its performance better, but it is an important part of formulation of a highly reliable system.

\subsubsection{Ethical and Legal Requirement}

A first requirement is to avoid algorithmic discrimination.
Due to the nature of machine learning techniques, a trained deep neural network may inherit the bias in the training set, which is sometimes hard to notice.
There is a concern of fairness when DNNs are used in our daily life, for instance, mortgage qualification, credit and insurance risk assessments.

Deep neural networks have also been used for new drug discovery and design~\cite{2016dlidd}.
The computational drug design field was dominated by conventional machine learning methods such as random forests and generalized additive models, partially because of their efficient learning algorithms at that time, and also because a domain chemical interpretation is possible.
Interpretability is also needed for a new drug to get approved by the regulator, such as the Food and Drug Administration (FDA).
Besides the clinical test results, the biological mechanism underpinning the results is usually required.
The same goes for medical devices.

Another legal requirement of interpretability is the ``right to explanation''~\cite{2017euroa}.
According to the EU General Data Protection Regulation (GDPR)~\cite{2016gdpr}, Article 22, people have the right not to be subject to an automated decision which would produce legal effects or similar significant effects concerning him or her.
The data controller shall safeguard the data owner's right to obtain human intervention, to express his or her point of view and to contest the decision.
If we have no idea how the network makes a decision, there is no way to ensure these rights.

\subsubsection{Scientific Usage}

Deep neural networks are becoming powerful tools in scientific research fields where the data may have complex intrinsic patterns (e.g., genomics~\cite{2015dlfrg}, astronomy~\cite{parks2018dloqs}, physics~\cite{2014sfepi} and even social science~\cite{2017paeis}).
The word ``science'' is derived from the Latin word ``scientia'', which means ``knowledge''.
When deep networks reach a better performance than the old models, they must have found some unknown ``knowledge''.
Interpretability is a way to reveal it.

\subsection{Related Work and Contributions}

There have already been attempts to summarize the techniques for neural network interpretability.
However, most of them only provide basic categorization or enumeration, without a clear taxonomy.
Lipton~\cite{lipton2016tmomi} points out that the term interpretability is not well-defined and often has different meanings in different studies.
He then provides simple categorization of both the need (e.g., trust, causality, fair decision-making etc.) and methods (post-hoc explanations) in interpretability study.
Doshi-Velez and Kim~\cite{doshi2017tarso} provide a discussion on the definition and evaluation of interpretability, which inspired us to formulate a stricter definition and to categorize the existing methods based on it.
Montavon et al.~\cite{montavon2018mfiau} confine the definition of explanation to feature importance (also called explanation vectors elsewhere) and review the techniques to interpret learned concepts and individual predictions by networks.
They do not aim to give a comprehensive overview and only include some representative approaches.
Gilpin et al.~\cite{gilpin2018eeaoo} divide the approaches into three categories: explaining data processing, explaining data representation and explanation-producing networks.
Under this categorization, the linear proxy model method and the rule-extraction method are equally viewed as proxy methods, without noticing many differences between them (the former is a local method while the latter is usually global and their produced explanations are different, we will see it in our taxonomy).
Guidotti et al.~\cite{guidotti2018asomf} consider all black-box models (including tree ensembles, SVMs etc.) and give a fine-grained classification based on four dimensions (the type of interpretability problem, the type of explanator, the type of black-box model, and the type of data).
However, they treat decision trees, decision rules, saliency masks, sensitivity analysis, activation maximization etc.\ equally, as explanators.
    In our view, some of them are certain types of explanations while some of them are methods used to produce explanations.
Zhang and Zhu~\cite{2018vifdl} review the methods to understand network's mid-layer representations or to learn networks with interpretable representations in computer vision field.

This survey has the following contributions:

\begin{itemize}
    %% stricter/practical definition
    \item We make a further step towards the definition of interpretability on the basis of reference~\cite{doshi2017tarso}.
          In this definition, we emphasize the \textit{type} (or \textit{format}) \textit{of explanations} (e.g, rule forms, including both decision trees and decision rule sets).
          This acts as an important dimension in our proposed taxonomy.
          Previous papers usually organize existing methods into various isolated (to a large extent) \textit{explanators} (e.g., decision trees, decision rules, feature importance, saliency maps etc.).
    \item We analyse the real needs for interpretability and summarize them into 3 groups: interpretability as an important component in systems that should be highly-reliable, ethical or legal requirements, and interpretability providing tools to enhance knowledge in the relevant science fields.
          In contrast, a previous survey~\cite{guidotti2018asomf} only shows the importance of interpretability by providing several cases where black-box models can be dangerous.
    \item We propose a new taxonomy comprising three dimensions (passive vs.\ active approaches, the format of explanations, and local-semilocal-global interpretability).
          Note that although many ingredients of the taxonomy have been discussed in the previous literature, they were either mentioned in totally different context, or entangled with each other.
          To the best of our knowledge, our taxonomy provides the most comprehensive and clear categorization of the existing approaches.
\end{itemize}

The three degrees of freedom along which our taxonomy is organized allow for a schematic 3D view illustrating how diverse attempts at interpretability of deep networks are related.
It also provides suggestions for possible future work by filling some of the gaps in the interpretability research (see Figure~\ref{fig:paper-space}).

\subsection{Organization of the Survey}\label{sec:organization}

The rest of the survey is organized as follows.
In Section~\ref{sec:taxonomy}, we introduce our proposed taxonomy for network interpretation methods.
The taxonomy consists of three dimensions, passive vs.\ active methods, type of explanations and global vs.\ local interpretability.
Along the first dimension, we divide the methods into two groups, passive methods (Section~\ref{sec:passive}) and active methods (Section~\ref{sec:active}).
Under each section, we {traverse the remaining two dimensions (different kinds of explanations, and whether they are local, semi-local or global).
Section~\ref{sec:eval} gives a brief summary of the evaluation of interpretability.}
Finally, we conclude this survey in Section~\ref{sec:conclusion}.

\section{Taxonomy}\label{sec:taxonomy}

\begin{figure*}[!t]
    \centering
    \input{imgs/taxonomy.tikz}
    \caption{The 3 dimensions of our taxonomy.}\label{fig:dimensions}
\end{figure*}

We propose a novel taxonomy with three dimensions (see Figure~\ref{fig:dimensions}): (1) the passive vs.\ active approaches dimension, (2) the type/format of produced explanations, and (3) from local to global interpretability dimension respectively.
\textbf{The first dimension} is categorical and has two possible values, \textit{passive interpretation} and \textit{active interpretability intervention}.
It divides the existing approaches according to whether they require to change the network architecture or the optimization process.
The passive interpretation process starts from a trained network, with all the weights already learned from the training set.
Thereafter, the methods try to extract logic rules or extract some understandable patterns.
In contrast, active methods require some changes before the training, such as introducing extra network structures or modifying the training process.
These modifications encourage the network to become more interpretable (e.g., more like a decision tree).
Most commonly such active interventions come in the form of regularization terms.

In contrast to previous surveys, the other two dimensions allow ordinal values.
For example, the previously proposed dimension \textit{type of explanator}~\cite{guidotti2018asomf} produces subcategories like \textit{decision trees}, \textit{decision rules}, \textit{feature importance}, \textit{sensitivity analysis} etc.
However, there is no clear connection among these pre-recognised explanators (what is the relation between decision trees and feature importance).
Instead, our \textbf{second dimension} is \textit{type/format of explanation}.
By inspecting various kinds of explanations produced by different approaches, we can observe differences in how explicit they are.
Logic rules provide the most clear and explicit explanations while other kinds of explanations may be implicit.
For example, a saliency map itself is just a mask on top of a certain input.
By looking at the saliency map, people construct an explanation ``the model made this prediction because it focused on this highly influential part and that part (of the input)''.
Hopefully, these parts correspond to some domain understandable concepts.
Strictly speaking, implicit explanations by themselves are not complete explanations and need further human interpretation, which is usually automatically done when people see them.
We recognize four major types of explanations here, \textit{logic rules}, \textit{hidden semantics}, \textit{attribution} and \textit{explanations by examples}, listed in order of decreasing {explanatory power}.
Similar discussions can be found in the previous literature, e.g., Samek et~al.~\cite{samek2019teai} provide a short subsection about ``type of explanations'' (including explaining learned representations, explaining \textit{individual predictions} etc.).
However, it is mixed up with another independent dimension of the interpretability research which we will introduce in the following paragraph.
A recent survey~\cite{bodria2021benchmarking} follows the same philosophy
and treats saliency maps and concept attribution~\cite{kim2018ibfaq} as different types of explanations,
while we view them as being of the same kind, but differing in the dimension below.

\textbf{The last dimension}, from local to global interpretability (w.r.t.\ the input space), has become very common in recent papers (e.g.,~\cite{doshi2017tarso,guidotti2018asomf,montavon2018mfiau},\cite{molnar2020interpretable}),
where global interpretability means being able to understand the overall decision logic of a model and local interpretability focuses on the explanations of individual predictions.
However, in our proposed dimension, there exists a transition rather than a hard division between global and local interpretability (i.e.\ semi-local interpretability).
Local explanations usually make use of the information at the target input (e.g., its feature values, its gradient).
But global explanations try to generalize to as wide ranges of inputs as possible (e.g., sequential \textit{covering} in rule learning, \textit{marginal} contribution for feature importance ranking).
This view is also supported by the existence of several semi-local explanation methods~\cite{ribeiro2018ahpma,adler2018abbmf}.
There have also been attempts to fuse local explanations into global ones in a bottom-up fashion~\cite{pedreschi2019meobb,lapuschkin2019uchpa,natesan2020mame}.

To help understand the latter two dimensions, Table~\ref{tab:example-explanation} lists examples of typical explanations produced by different subcategories under our taxonomy.
{\bf (Row 1)} When considering \textit{rule as explanation for local interpretability}, an example is to provide rule explanations which only apply to a given input \(\bm{x}^{(i)}\) (and its associated output \(\hat{y}^{(i)}\)).
One of the solutions is to find out (by perturbing the input features and seeing how the output changes) the minimal set of features  \(\bm{x}_k{\dots} \bm{x}_l\) whose presence supports the prediction \(\hat{y}^{(i)}\).
Analogously, features \(\bm{x}_m{\dots} \bm{x}_n\) can be found which should not be present (larger values), otherwise \(\hat{y}^{(i)}\) will change.
Then an explanation rule for \(\bm{x}^{(i)}\) can be constructed as ``it is because \(\bm{x}_k{\dots} \bm{x}_l\) are present and \(\bm{x}_m{\dots} \bm{x}_n\) are absent that \(\bm{x}^{(i)}\) is classified as \(\hat{y}^{(i)}\)''~\cite{dhurandhar2018ebotm}.
If a rule is valid not only for the input \(\bm{x}^{(i)}\), but also for its ``neighbourhood''~\cite{ribeiro2018ahpma}, we obtain a \textit{semi-local interpretability}.
And if a rule set or decision tree is extracted from the original network, it explains the general function of the whole network and thus provides \textit{global interpretability}.
{\bf (Row 2)} When it comes to \textit{explaining the hidden semantics}, a typical example (\textit{global}) is to visualize what pattern a hidden neuron is mostly sensitive to.
This can then provide clues about the inner workings of the network.
We can also take a more pro-active approach to make hidden neurons more interpretable.
As a high-layer hidden neuron may learn a mixture of patterns that can be hard to interpret, Zhang et al.~\cite{zhang2018icnn} introduced a loss term that makes high-layer filters either produce consistent activation maps (among different inputs) or keep inactive (when not seeing a certain pattern).
Experiments show that those filters are more interpretable (e.g., a filter may be found to be activated by the head parts of animals).
{\bf (Row 3)} \textit{Attribution as explanation} usually provides \textit{local interpretability}.
Thinking about an animal classification task, input features are all the pixels of the input image.
Attribution allows people to see which regions (pixels) of the image contribute the most to the classification result.
The attribution can be computed e.g., by sensitivity analysis in terms of the input features (i.e.\ all pixels) or some variants~\cite{simonyan2013dicnv,zeiler2014vaucn}.
For \textit{attribution for global interpretability}, deep neural networks usually cannot have as straightforward attribution as e.g., coefficients \(\bm{w}\) in linear models \(y=\bm{w}^{\top}\bm{x}+b\), which directly show the importance of features globally.
Instead of concentrating on input features (pixels), Kim et al.~\cite{kim2018ibfaq} were interested in attribution to a ``concept'' (e.g., how sensitive is a prediction of zebra to the presence of stripes).
The concept (stripes) is represented by the normal vector to the plane which separates having-stripes and non-stripes training examples in the space of network's hidden layer.
It is therefore possible to compute how sensitive the prediction (of zebra) is to the concept (presence of stripes) and thus have some form of global interpretability.
{\bf (Row 4)} Sometimes researchers explain network prediction by \textit{showing other known examples} providing similar network functionality.
To explain a single input \(\bm{x}^{(i)}\) (\textit{local interpretability}), we can find an example which is most similar to \(\bm{x}^{(i)}\) in the network's hidden layer level.
This selection of explanation examples can also be done by testing how much the prediction of \(\bm{x}^{(i)}\) will be affected if a certain example is removed from the training set~\cite{koh2017ubbpv}.
To provide \textit{global interpretability by showing examples}, a method is adding a (learnable) prototype layer to a network.
The prototype layer forces the network to make predictions according to the proximity between input and the learned prototypes.
Those learned and interpretable prototypes can help to explain the network's overall function.

%% Reduce vertical spacing of math display mode
% \the\abovedisplayskip
% \the\belowdisplayskip
% \the\abovedisplayshortskip
% \the\belowdisplayshortskip
%% 6.7082pt + 4.015pt - 2.00749pt
\setlength{\abovedisplayskip}{2pt}
\setlength{\belowdisplayskip}{0pt}
\setlength{\abovedisplayshortskip}{2pt}
\setlength{\belowdisplayshortskip}{0pt}

\begin{table*}[!t]
    \centering
    \begin{threeparttable}
        \footnotesize
        \renewcommand{\arraystretch}{1.5}
        \setlength{\fboxsep}{2pt}
        \caption{Example explanations of networks. \upshape Please see Section~\ref{sec:taxonomy} for details. Due to lack of space, we do not provide examples for semi-local interpretability here. (We thank the anonymous reviewer for the idea to improve the clarity of this table.)}.\label{tab:example-explanation}
        \begin{tabular}{|c|m{0.38\textwidth}|m{0.38\textwidth}|}
            \hline
            & \multicolumn{1}{>{\centering\arraybackslash}m{0.38\textwidth}|}{\textbf{Local (and semi-local) interpretability}\par applies to a certain input \(\bm{x}^{(i)}\) (and its associated output \(\hat{y}^{(i)}\)), or a small range of inputs-outputs} & \multicolumn{1}{>{\centering\arraybackslash}m{0.38\textwidth}|}{\textbf{Global interpretability}\par w.r.t.\ the whole input space} \\
            \hline
            \multicolumn{1}{|>{\centering\arraybackslash}m{0.15\textwidth}|}{\textbf{Rule as explanation}}
            & \begin{center}\colorbox{gray!20}{Explain a certain \((\bm{x}^{(i)},y^{(i)})\) with a decision rule:}\end{center}\par\vspace{3pt}\textbullet{} The result ``\(\bm{x}^{(i)}\) is classified as \(\hat{y}^{(i)}\)'' is because \(\bm{x}_1,\bm{x}_4,\dots\) are present and \(\bm{x}_3,\bm{x}_5,\dots\) are absent~\cite{dhurandhar2018ebotm}.\par\vspace{3pt}\textbullet{} (Semi-local) For \(\bm{x}\) in the neighbourhood of \(\bm{x}^{(i)}\), if \((\bm{x}_1 > \alpha) \land (\bm{x}_3 < \beta) \land \dots\), then \(y = \hat{y}^{(i)}\)~\cite{ribeiro2018ahpma}.
            & \vspace{3.5pt}\begin{center}\colorbox{gray!20}{Explain the whole model \(y(\bm{x})\) with a decision rule set:}\end{center}\par\vspace{3pt}The neural network can be approximated by\[\begin{cases}
                \text{If }(\bm{x}_2 < \alpha) \land (\bm{x}_3 > \beta) \land \dots & \text{, then }y = 1,\\
                \text{If }(\bm{x}_1 > \gamma) \land (\bm{x}_5 < \delta) \land \dots & \text{, then }y = 2,\\
                \cdots & \\
                \text{If }(\bm{x}_4 \dots) \land (\bm{x}_7 \dots) \land \dots & \text{, then }y = M
            \end{cases}\]\\
            \hline
            \multicolumn{1}{|>{\centering\arraybackslash}m{0.15\textwidth}|}{\textbf{Explaining}\par\textbf{hidden semantics}\par{\scriptsize (make sense of certain}\par{\scriptsize hidden neurons/layers)}}
            & \begin{center}\colorbox{gray!20}{Explain a hidden neuron/layer \(h(\bm{x}^{(i)})\):}\end{center}\par\vspace{3.5pt}(*No explicit methods but many local attribution methods (see below) can be easily modified to ``explain'' a hidden neuron \(h(\bm{x})\) rather than the final output \(y\)).
            & \vspace{3.5pt}\begin{center}\colorbox{gray!20}{Explain a hidden neuron/layer \(h(\bm{x})\) instead of \(y(\bm{x})\):}\end{center}\par\vspace{3.5pt}\input{imgs/cell22.tikz} \\
            \hline
            \multicolumn{1}{|>{\centering\arraybackslash}m{0.15\textwidth}|}{\textbf{Attribution}\par \textbf{as explanation}}
            & \vspace{3.5pt}\begin{center}\colorbox{gray!20}{Explain a certain \((\bm{x}^{(i)},y^{(i)})\) with an attribution \(\bm{a}^{(i)}\):}\end{center}\par\vspace{3.5pt}\input{imgs/cell31.tikz}
            & \begin{center}\colorbox{gray!20}{Explain \(y(\bm{x})\) with attribution to certain features in general:}\end{center}\par\vspace{3.5pt}(Note that for a linear model, the coefficients is the global attribution to its input features.)\par\vspace{4pt}\textbullet{} Kim et al.~\cite{kim2018ibfaq} calculate attribution to a target ``concept'' rather than the input pixels of a certain input. For example, ``how sensitive is the output (a prediction of zebra) to a concept (the presence of stripes)?''\\
            \hline
            \multicolumn{1}{|>{\centering\arraybackslash}m{0.15\textwidth}|}{\textbf{Explanation by showing examples}}
            & \vspace{3.5pt}\begin{center}\colorbox{gray!20}{Explain a certain \((\bm{x}^{(i)},y^{(i)})\) with another \(\bm{x}^{(i)\prime}\):}\end{center}\par\vspace{3.5pt}\input{imgs/cell41.tikz}
            & \begin{center}\colorbox{gray!20}{Explain \(y(\bm{x})\) collectively with a few prototypes:}\end{center}\par\vspace{3.5pt}\textbullet{} Adds a (learnable) prototype layer to the network.\par Every prototype should be similar to at least an encoded input. Every input should be similar to at least a prototype. The trained network explains itself by its prototypes.~\cite{li2018dlfcb}\\
            \hline
        \end{tabular}
        \begin{tablenotes}
            \item [1] the contribution to the network prediction of \(\bm{x}^{(i)}\).
            \item [2] how sensitive is the classification result to the change of pixels.
            \item [3] without the training image, the network prediction of \(\bm{x}^{(i)}\) will change a lot. In other words, these images help the network make a decision on \(\bm{x}^{(i)}\).
        \end{tablenotes}
    \end{threeparttable}
\end{table*}

%% Restore
\setlength{\abovedisplayskip}{8.716pt}
\setlength{\belowdisplayskip}{8.716pt}
\setlength{\abovedisplayshortskip}{8.716pt}
\setlength{\belowdisplayshortskip}{8.716pt}

\begin{figure}[!t]
    \centering
    \includegraphics[width=\columnwidth]{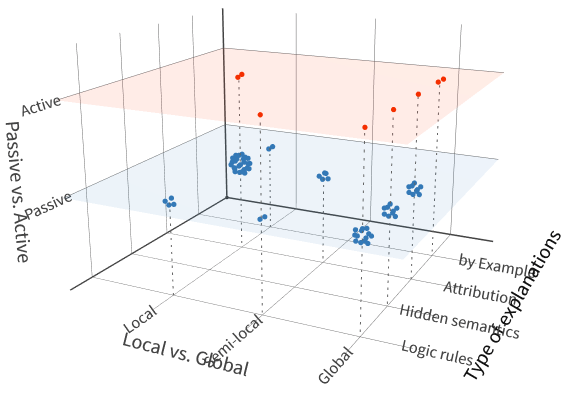}
    \caption{The distribution of the interpretability papers in the 3D space of our taxonomy. We can rotate and observe the density of work in certain areas/planes and find the missing parts of interpretability research. (See \url{https://yzhang-gh.github.io/tmp-data/index.html})}\label{fig:paper-space}
\end{figure}

With the three dimensions introduced above, we can visualize the distribution of the existing interpretability papers in a 3D view (Figure~\ref{fig:paper-space} only provides a 2D snapshot, we encourage readers to visit the online interactive version for better presentation).
Table~\ref{tab:papers} is another representation of all the reviewed interpretability approaches which is good for quick navigation.

\begin{table*}
    \centering
    \begin{threeparttable}
    \footnotesize
    \renewcommand{\arraystretch}{1.5}
    \caption{An overview of the interpretability papers.}\label{tab:papers}
    \begin{tabular}{|l|l|m{0.22\textwidth}|m{0.16\textwidth}|m{0.24\textwidth}|}
        \hline
                                                                      &                                       & \tableheadwithvline{Local}                     & \tableheadwithvline{Semi-local}                & \tableheadwithvline{Global}              \\
        \hline\hline
        \multicolumn{1}{|c|}{\bfseries\multirow{4}{*}[-3em]{Passive}} & \tableheadwithvline{Rule}             & CEM{\supcite{dhurandhar2018ebotm}},
                                                                                                                CDRPs{\supcite{2018innbi}},
                                                                                                                CVE\textsuperscript{2}{\supcite{goyal2019cve}},\newline
                                                                                                                DACE{\supcite{kanamori2020dace}}               & Anchors{\supcite{ribeiro2018ahpma}},\newline
                                                                                                                                                                 Interpretable partial~substitution{\supcite{wang2019gfolc}}
                                                                                                                                                                                                                & KT{\supcite{fu1991rlbso}},
                                                                                                                                                                                                                  {\(M\)of\(N\)\supcite{1993errfk}},
                                                                                                                                                                                                                  NeuralRule{\supcite{1995unnvr}},
                                                                                                                                                                                                                  NeuroLinear{\supcite{1997nfnnt}},
                                                                                                                                                                                                                  GRG{\supcite{odajima2008grgfd}},\newline
                                                                                                                                                                                                                  Gyan\textsuperscript{FO}{\supcite{nayak2009grwpt}},
                                                                                                                                                                                                                  \textbullet\textsuperscript{FZ}{\supcite{1997aannb,castro2002ioann}},
                                                                                                                                                                                                                  Trepan{\supcite{craven1996etsro}},
                                                                                                                                                                                                                  \textbullet{\supcite{krishnan1999edtft}},
                                                                                                                                                                                                                  DecText{\supcite{boz2002edtft}}, Global model on CEM{\supcite{pedapati2020lgtmc}} \\
        \cline{2-5}
                                                                      & \tableheadwithvline{Hidden semantics} & (*No explicit methods but many in the below cell can be applied here.) & \multicolumn{1}{c|}{---}                 & Visualization\supcite{2009vhfoa,simonyan2013dicnv,wang2018vdnnb,mahendran2015udirb,yosinski2015unntd,nguyen2016stpif,2017fv},\newline
                                                                                                                                                                                                                  Network~dissection\supcite{bau2017ndqio},
                                                                                                                                                                                                                  Net2Vec\supcite{fong2018nqaeh},\newline    
                                                                                                                                                                                                                  Linguistic correlation analysis\supcite{dalvi2019wiogo}  \\
        \cline{2-5}
                                                                      & \tableheadwithvline{Attribution\textsuperscript{1}}      & LIME{\supcite{ribeiro2016wsity}},
                                                                                                                MAPLE{\supcite{plumb2018masle}},
                                                                                                                Partial derivatives{\supcite{simonyan2013dicnv}},
                                                                                                                DeconvNet{\supcite{zeiler2014vaucn}},\newline
                                                                                                                Guided~backprop{\supcite{springenberg2014sfsta}},
                                                                                                                Guided Grad-CAM{\supcite{selvaraju2017gcvef}},
                                                                                                                Shapley~values{\supcite{strumbelj2010aeeoi,lundberg2017auati,ancona2019ednnw,heskes2020csvec}},\newline
                                                                                                                Sensitivity analysis{\supcite{zeiler2014vaucn,fong2017ieobb,zintgraf2017vdnnd}},\newline
                                                                                                                Feature~selector{\supcite{chen2018lteai}},\newline
                                                                                                                Bias attribution{\supcite{wang2019bamba}}
                                                                                                                                                               & DeepLIFT{\supcite{shrikumar2017liftp}},
                                                                                                                                                                 LRP{\supcite{bach2015opwef}},\newline
                                                                                                                                                                 Integrated~gradients{\supcite{sundararajan2017aafdn}},\newline
                                                                                                                                                                 Feature~selector{\supcite{chen2018lteai}},\newline MAME{\supcite{natesan2020mame}}      & Feature~selector{\supcite{chen2018lteai}},
                                                                                                                                                                                                                  TCAV{\supcite{kim2018ibfaq}},\newline ACE{\supcite{ghorbani2019tacbe}}, SpRAy\textsuperscript{3}{\supcite{lapuschkin2019uchpa}}, MAME{\supcite{natesan2020mame}}\newline DeepConsensus{\supcite{salman2020deep}} \\
        \cline{2-5}
                                                                      & \tableheadwithvline{By example}       & Influence~functions{\supcite{koh2017ubbpv}},\newline
                                                                                                                Representer~point~selection{\supcite{yeh2018rpsfe}} & \multicolumn{1}{c|}{---}                  & \multicolumn{1}{c|}{---}                   \\
        \hline\hline
        \multicolumn{1}{|c|}{\bfseries\multirow{4}{*}[-0.5em]{Active}} & \tableheadwithvline{Rule}            & \multicolumn{1}{c|}{---} & Regional tree\newline regularization{\supcite{wu2020rtrfi}} & Tree regularization{\supcite{wu2018bstro}}    \\
        \cline{2-5}
                                                               & \tableheadwithvline{Hidden semantics} & \multicolumn{1}{c|}{---} & \multicolumn{1}{c|}{---} & ``One filter, one concept''{\supcite{zhang2018icnn}}  \\
        \cline{2-5}
                                                               & \tableheadwithvline{Attribution}      & ExpO{\supcite{plumb2020rbbmf}}, DAPr{\supcite{weinberger2020ldapb}} & \multicolumn{1}{c|}{---} & Dual-net (feature importance){\supcite{wojtas2020firfd}}                              \\
        \cline{2-5}
                                                               & \tableheadwithvline{By example}       & \multicolumn{1}{c|}{---} & \multicolumn{1}{c|}{---} & Network with a prototype layer{\supcite{li2018dlfcb}},\newline
                                                                                                                                                               ProtoPNet{\supcite{chen2019tlltd}} \\
        \hline
    \end{tabular}
    \begin{tablenotes}
        \item [FO] First-order rule
        \item [FZ] Fuzzy rule
        \item [1] Some attribution methods (e.g., DeconvNet, Guided Backprop) arguably have certain non-locality because of the rectification operation.
        \item [2] Short for counterfactual visual explanations
        \item [3] SpRAy is flexible to provide semi-local or global explanations by clustering local (individual) attributions.
    \end{tablenotes}
    \end{threeparttable}
\end{table*}

In the following the sections, we will scan through Table~\ref{tab:papers} along each dimension.
The first dimension results in two sections, passive methods (Section~\ref{sec:passive}) and active methods (Section~\ref{sec:active}).
We then expand each section to several subsections according to the second dimension (type of explanation).
Under each subsection, we introduce (semi-)local vs.\ global interpretability methods respectively.

\section{Passive Interpretation of Trained Networks}\label{sec:passive}

Most of the existing network interpreting methods are passive methods.
They try to understand the already trained networks.
We now introduce these methods according to their types of produced explanations (i.e.\ the second dimension).

\subsection{Passive, \textbf{Rule as Explanation}}

Logic rules are commonly acknowledged to be interpretable and have a long history of research.
Thus rule extraction is an appealing approach to interpret neural networks.
In most cases, rule extraction methods provide \textit{global} explanations as they only extract a single rule set or decision tree from the target model.
There are only a few methods producing \textit{(semi-)local} rule-form explanations which we will introduce below (Section~\ref{sec:passive-rule-local}), followed are global methods (Section~\ref{sec:passive-rule-global}).
Another thing to note is that although the rules and decision trees (and their extraction methods) can be quite different, we do not explicitly differentiate them here as they provide similar explanations (a decision tree can be flattened to a decision rule set).
A basic form of a rule is
\begin{equation*}
    \text{If } P \text{, then } Q.
\end{equation*}
where \(P\) is called the antecedent, and \(Q\) is called the consequent, which in our context is the prediction (e.g., class label) of a network.
\(P\) is usually a combination of conditions on several input features.
For complex models, the explanation rules can be of other forms such as the propositional rule, first-order rule or fuzzy rule.

\subsubsection{Passive, Rule as Explanation, \textbf{(Semi-)local}}\label{sec:passive-rule-local}

According to our taxonomy, methods in this category focus on a trained neural network and a certain input (or a small group of inputs), and produce a logic rule as an explanation.
Dhurandhar et al.~\cite{dhurandhar2018ebotm} construct local rule explanations by finding out features that should be minimally and sufficiently \textit{present} and features that should be minimally and necessarily \textit{absent}.
In short, the explanation takes this form ``\textit{If an input \(\bm{x}\) is classified as class \(y\), it is because features \(f_i, \dots, f_k\) are present and features \(f_m, \dots, f_p\) are absent}''.
This is done by finding small sparse perturbations that are sufficient to ensure the same prediction by its own (or will change the prediction if applied to a target input)\footnote{The authors also extended this method to learn a \textit{global} interpretable model, e.g., a decision tree, based on custom features created from above \textit{local} explanations~\cite{pedapati2020lgtmc}.}.
A similar kind of methods is counterfactual explanations~\cite{wachter2018counterfactual}.
Usually, we are asking based on what features ('s values) the neural network makes the prediction of class \(c\).
However, Goyal et al.~\cite{goyal2019cve} try to find the minimum-edit on an input image which can result in a different predicted class \(c'\).
In other words, they ask: ``What region in the input image makes the prediction to be class \(c\), \textit{rather than \(c'\)}''.
Kanamori et al.~\cite{kanamori2020dace} introduced distribution-aware counterfactual explanations, which require above ``edit'' to follow the empirical data distribution instead of being arbitrary.

Wang et al.~\cite{2018innbi} came up with another local interpretability method, which identifies \textit{critical data routing paths} (CDRPs) of the network for each input.
In convolutional neural networks, each kernel produces a feature map that will be fed into the next layer as a channel.
Wang et al.~\cite{2018innbi} associated every output channel on each layer with a gate (non-negative weight), which indicates how critical that channel is.
These gate weights are then optimized such that when they are multiplied with the corresponding output channels, the network can still make the same prediction as the original network (on a given input).
Importantly, the weights are encouraged to be sparse (most are close to zero).
CDRPs can then be identified for each input by first identifying the \textit{critical nodes}, i.e.\ the intermediate kernels associated with positive gates.
We can explore and assign meanings to the critical nodes so that the critical paths become local explanations.
However, as the original paper did not go further on the CDRPs representation which may not be human-understandable, it is still more of an activation pattern than a real explanation.

We can also extract rules that cover a group of inputs rather than a single one.
Ribeiro et al.~\cite{ribeiro2018ahpma} propose \textit{anchors} which are if-then rules that are sufficiently precise (semi-)locally.
In other words, if a rule applies to a group of similar examples, their predictions are (almost) always the same.
It is similar to (actually, on the basis of) an attribution method LIME, which we will introduce in Section~\ref{sec:attr}.
However, they are different in terms of the produced explanations (LIME produces attribution for individual examples).
Wang et al.~\cite{wang2019gfolc} attempted to find an interpretable partial substitution (a rule set) to the network that covers a certain subset of the input space.
This substitution can be done with no or low cost on the model accuracy according to the size of the subset.

\subsubsection{Passive, Rule as Explanation, \textbf{Global}}\label{sec:passive-rule-global}

Most of the time, we would like to have some form of an overall interpretation of the network, rather than its local behaviour at a single point.
We again divide these approaches into two groups.
Some rule extraction methods make use of the network-specific information such as the network structure, or the learned weights.
These methods are called \textit{decompositional} approaches in previous literature~\cite{craven1994usaqt}.
The other methods instead view the network as a black-box and only use it to generate training examples for classic rule learning algorithms.
They are called \textit{pedagogical} approaches.

\paragraph{Decompositional approaches}

Decompositional approaches generate rules by observing the connections in a network.
As many of the these approaches were developed before the deep learning era, they are mostly designed for classic fully-connected feedforward networks.
Considering a single-layer setting of a fully-connected network (only one output neuron),
\begin{equation*}
    y = \sigma\rbracket{\sum_i \bm{w}_i \bm{x}_i + b}
\end{equation*}
where \( \sigma \) is an activation function (usually sigmoid, \(\sigma(x)=1/(1+\mathrm{e}^{-x})\)), \(\bm{w}\) are the trainable weights, \(\bm{x}\) is the input vector, and \(b\) is the bias term (often referred as a threshold \( \theta \) is the early time, and \(b\) here can be interpreted as the negation of \( \theta \)).
Lying at the heart of rule extraction is to search for combinations of certain values (or ranges) of attributes \(\bm{x}_i\) that make \(y\) near \(1\)~\cite{1993errfk}.
This is tractable only when we are dealing with small networks because the size of the search space will soon grow to an astronomical number as the number of attributes and the possible values for each attribute increase.
Assuming we have \(n\) Boolean attributes \(\bm{x}_i\) as an input, and each attribute can be \texttt{\small true} or \texttt{\small false} or absent in the antecedent, there are \(\mathcal{O}(3^n)\) combinations to search.
We therefore need some search strategies.

One of the earliest methods is the KT algorithm~\cite{fu1991rlbso}.
KT algorithm first divides the input attributes into two groups, pos-atts (short for positive attributes) and neg-atts, according to the signs of their corresponding weights.
Assuming activation function is sigmoid, all the neurons are booleanized to \texttt{\small true} (if close enough to \(1\)) or \texttt{\small false} (close to \(0\)).
Then, all combinations of pos-atts are selected if the combination can on its own make \(y\) be \texttt{\small true} (larger than a pre-defined threshold \( \beta \) without considering the neg-atts), for instance, a combination \( \{\bm{x}_1, \bm{x}_3\} \) and \(\sigma(\sum_{i \in \{1,3\}} \bm{w}_i \bm{x}_i + b) > \beta\).
Finally, it takes into account the neg-atts.
For each above pos-atts combination, it finds combinations of neg-atts (e.g., \( \{\bm{x}_2, \bm{x}_5\} \)) that when absent the output calculated from the selected pos-atts and unselected neg-atts is still \texttt{\small true}.
In other words, \(\sigma(\sum_{i \in \mathcal{I}} \bm{w}_i \bm{x}_i + b) > \beta\), where \(\mathcal{I} = \{\bm{x}_1, \bm{x}_3\} \cup \{\text{neg-atts}\} \setminus \{\bm{x}_2, \bm{x}_5\}\).
The extracted rule can then be formed from the combination \(\mathcal{I}\) and has the output class \(1\).
In our example, the translated rule is
\begin{equation*}
    \text{If } \bm{x}_1 \text{(is \texttt{\small true})} \land \bm{x}_3 \land \neg \bm{x}_2 \land \neg \bm{x}_5, \text{then } y=1.
\end{equation*}
Similarly, this algorithm can generate rules for class \(0\) (searching neg-atts first and then adding pos-atts).
To apply to the multi-layer network situation, it first does layer-by-layer rule generation and then rewrites them to omit the hidden neurons.
In terms of the complexity, KT algorithm reduces the search space to \(\mathcal{O}(2^n)\) by distinguishing pos-atts and neg-atts (pos-atts will be either \texttt{\small true} or absent, and neg-atts will be either \texttt{\small false} or absent).
It also limits the number of attributes in the antecedent, which can further decrease the algorithm complexity (with the risk of missing some rules).

Towell and Shavlik~\cite{1993errfk} focus on another kind of rules of ``\(M\)-of-\(N\)'' style.
This kind of rule de-emphasizes the individual importance of input attributes, which has the form
\begin{equation*}
    \text{If } M \text{ of these } N \text{ expressions are } \texttt{\small true}, \text{then } Q.
\end{equation*}
This algorithm has two salient characteristics.
The first one is link (weight) clustering and reassigning them the average weight within the cluster.
Another characteristic is network simplifying (eliminating unimportant clusters) and re-training.
Comparing with the exponential complexity of subset searching algorithm, \(M\)-of-\(N\) method is approximately cubic because of its special rule form.

NeuroRule~\cite{1995unnvr} introduced a three-step procedure of extracting rules:
(1) train the network and prune,
(2) discretize (cluster) the activation values of the hidden neurons,
(3) extract the rules layer-wise and rewrite (similar as previous methods).
NeuroLinear~\cite{1997nfnnt} made a little change to the NeuroRule method, allowing neural networks to have continuous input.
Andrews et al.~\cite{1995sacot} and Tickle et al.~\cite{tickle1998ttwct} provide a good summary of the rule extraction techniques before 1998.

\paragraph{Pedagogical approaches}

By treating the neural network as a black-box, \textit{pedagogical} methods (or hybrids of both) directly learn rules from the examples generated by the network.
It is essentially reduced to a traditional rule learning or decision tree learning problem.
For rule set learning, we have sequential covering framework (i.e.\ to learn rules one by one).
And for decision tree, there are many classic algorithms like CART~\cite{1984cart} and C4.5~\cite{1993cpfml}.
Example work of decision tree extraction (from neural networks) can be found in references~\cite{craven1996etsro,krishnan1999edtft,boz2002edtft}.

Odajima et al.~\cite{odajima2008grgfd} followed the framework of NeuroLinear but use a greedy form of sequential covering algorithm to extract rules.
It is reported to be able to extract more concise and precise rules.
Gyan method~\cite{nayak2009grwpt} goes further than extracting propositional rules.
After obtaining the propositional rules by the above methods, Gyan uses the Least General Generalization (LGG~\cite{1970anoig}) method to generate first-order logic rules from them.
There are also some approaches attempting to extract fuzzy logic from trained neural networks~\cite{1997aannb,2000nrgsi,castro2002ioann}.
The major difference is the introduction of the membership function of linguistic terms.
An example rule is
\begin{equation*}
    \text{If } (\bm{x}_1 = \texttt{\small high}) \land \dots, \text{then } y=\text{\texttt{\small class1}}.
\end{equation*}
where \texttt{\small high} is a fuzzy term expressed as a fuzzy set over the numbers.

Most of the above ``Rule as Explanation, Global'' methods were developed in the early stage of neural network research, and usually were only applied to relatively small datasets
(e.g., the Iris dataset, the Wine dataset from the UCI Machine Learning Repository).
However, as neural networks get deeper and deeper in recent applications, it is unlikely for a single decision tree to faithfully approximate the behaviour of deep networks.
We can see that more recent ``Rule as Explanation'' methods turn to local or semi-local interpretability~\cite{wang2019gfolc,wu2020rtrfi}.

\subsection{Passive, \textbf{Hidden Semantics as Explanation}}

The second typical kind of explanations is the meaning of hidden neurons or layers.
Similar to the grandmother cell hypothesis\footnote{\url{https://en.wikipedia.org/wiki/Grandmother_cell}} in neuroscience, it is driven by a desire to associate abstract concepts with the activation of some hidden neurons.
Taking animal classification as an example, some neurons may have high response to the head of an animal while others neurons may look for bodies, feet or other parts.
This kind of explanations by definition provides \textit{global} interpretability.

\subsubsection{Passive, Hidden Semantics as Explanation, \textbf{Global}}

Existing hidden semantics interpretation methods mainly focus on the computer vision field.
The most direct way is to show what the neuron is ``looking for'', i.e.\ visualization.
The key to visualization is to find a representative input that can maximize the activation of a certain neuron, channel or layer, which is usually called \textit{activation maximization}~\cite{2009vhfoa}.
This is an optimization problem, whose search space is the potentially huge input (sample) space.
Assuming we have a network taking as input a \(28\times28\) pixels black and white image (as in the MNIST handwritten digit dataset), there will be \(2^{28\times28}\) possible input images, although most of them are probably nonsensical.
In practice, although we can find a maximum activation input image with optimization, it will likely be unrealistic and uninterpretable.
This situation can be helped with some regularization techniques or priors.

We now give an overview over these techniques.
The framework of activation maximization was introduced by Erhan et al.~\cite{2009vhfoa} (although it was used in the unsupervised deep models like Deep Belief Networks).
In general, it can be formulated as
\[\mathbf{x}^{\star} = \argmax_{\mathbf{x}} (act(\mathbf{x};\theta) - \lambda \Omega(\mathbf{x}))\]
where \(act(\cdot)\) is the activation of the neuron of interest, \(\theta\) is the network parameters (weights and biases) and \(\Omega\) is an optional regularizer.
(We use the bold upright \(\bm{\mathrm{x}}\) to denote an input matrix in image related tasks, which allows row and column indices \(i\) and \(j\).)

Simonyan et al.~\cite{simonyan2013dicnv} for the first time applied activation maximization to a supervised deep convolutional network (for ImageNet classification).
It finds representative images by maximizing the score of a class (before softmax).
And the \(\Omega\) is the \(L_2\) norm of the image.
Later, people realized high frequency noise is a major nuisance that makes the visualization unrecognizable~\cite{2017fv,wang2018vdnnb}.
In order to get natural and useful visualizations, finding good priors or regularizers \(\Omega\) becomes the core task in this kind of approaches.

Mahendran and Vedaldi~\cite{mahendran2015udirb} propose a regularizer \textit{total variation}
\[\Omega(\mathbf{x}) = \sum_{i,j}{\left({(\mathbf{x}_{i,j+1}-\mathbf{x}_{i,j})}^2 + {(\mathbf{x}_{i+1,j}-\mathbf{x}_{i,j})}^2\right)}^{\frac{\beta}{2}}\]
which encourages neighbouring pixels to have similar values.
This can also be viewed as a low-pass filter that can reduce the high frequency noise in the image.
This kind of methods is usually called image blurring.
Besides suppressing high amplitude and high frequency information (with \(L_2\) decay and Gaussian blurring respectively), Yosinski et al.~\cite{yosinski2015unntd} also include other terms to clip the pixels of small values or little importance to the activation.
Instead of using many hand-crafted regularizers (image priors), Nguyen et al.~\cite{nguyen2016stpif} suggest using natural image prior learned by a generative model.
As Generative Adversarial Networks (GANs)~\cite{2014gan} showed recently great power to generate high-resolution realistic images~\cite{2016urlwd,2017psisu}, making use of the generative model of a GAN appears to be a good choice.
For a good summary and many impressive visualizations, we refer the readers to~\cite{2017fv}.
When applied to certain tasks, researchers can get some insights from these visual interpretations.
For example, Minematsu et al.~\cite{minematsu2017analytics,minematsu2018analytics} inspected the behaviour of the first and last layers of a DNN used for change detection (in video stream),
which may suggest the possibility of a new background modelling strategy.

Besides visualization, there are also some work trying to find connections between kernels and visual concepts (e.g., materials, certain objects).
Bau et al.~\cite{bau2017ndqio} (Network Dissection) collected a new dataset Broden which provides a pixel-wise binary mask \(L_c{(\mathbf{x})}\) for every concept \(c\) and each input image \(\mathbf{x}\).
The activation map of a kernel \(k\) is upscaled and converted (given a threshold) to a binary mask \(M_k(\mathbf{x})\) which has the same size of \(\mathbf{x}\).
Then the alignment between a kernel \(k\) and a certain concept \(c\) (e.g., car) is computed as
\[IoU_{k,c} = \frac{\sum|M_k(\mathbf{x}) \cap L_c(\mathbf{x})|}{\sum|M_k(\mathbf{x}) \cup L_c(\mathbf{x})|}\]
where \(|\cdot|\) is the cardinality of a set and the summation \(\sum\) is over all the inputs \(\mathbf{x}\) that contains the concept \(c\).
Along the same lines, Fong and Vedaldi~\cite{fong2018nqaeh} investigate the embeddings of concepts over multiple kernels by constructing \(M\) with a combination of several kernels.
Their experiments show that multiple kernels are usually required to encode one concept and kernel embeddings are better representations of the concepts.

Dalvi et al.~\cite{dalvi2019wiogo} also analysed the meaning of individual units/neurons in the networks for NLP tasks.
They build a linear model between the network's hidden neurons and the output.
The neurons are then ranked according to the significance of the weights of the linear model.
For those top-ranking neurons, their linguistic meanings are investigated by visualizing their saliency maps on the inputs, or by finding the top words by which they get activated.

\subsection{Passive, \textbf{Attribution as Explanation}}\label{sec:attr}

Attribution is to assign credit or blame to the input features in terms of their impact on the output (prediction).
The explanation will be a real-valued vector which indicates feature importance with the sign and amplitude of the scores~\cite{montavon2018mfiau}.
For simple models (e.g., linear models) with meaningful features, we might be able to assign each feature a score \textit{globally}.
When it comes to more complex networks and input, e.g., images, it is hard to say a certain pixel always has similar contribution to the output.
Thus, many methods do attribution \textit{locally}.
We introduce them below and at the end of this section we mention a global attribution method on intermediate representation rather than the original input features.

\subsubsection{Passive, Attribution as Explanation, \textbf{(Semi-)local}}

Similarly to the decompositional vs.\ pedagogical division of rule extraction methods, attribution methods can be also divided into two groups: gradient-related methods and model agnostic methods.

\paragraph{Gradient-related and backpropagation methods}\label{par:gradient-attribution}

Using gradients to explain the individual classification decisions is a natural idea as the gradient represents the ``direction'' and rate of the fastest increase on the loss function.
The gradients can also be computed with respect to a certain output class, for example, along which ``direction'' a perturbation will make an input more/less likely predicted as a cat/dog.
Baehrens et al.~\cite{baehrens2010hteic} use it to explain the predictions of Gaussian Process Classification (GPC), \(k\)-NN and SVM\@.
For a special case, the coefficients of features in linear models (or general additive models) are already the partial derivatives, in other words, the (global) attribution.
So people can directly know how the features affect the prediction and that is an important reason why linear models are commonly thought interpretable.
While plain gradients, discrete gradients and path-integrated gradients have been used for attribution, some other methods do not calculate real gradients with the chain rule but only backpropagate attribution signals (e.g., do extra normalization on each layer upon backpropagation).
We now introduce these methods in detail.

In computer vision, the attribution is usually represented as a saliency map, a mask of the same size of the input image.
In reference~\cite{simonyan2013dicnv}, the saliency map is generated from the gradients (specifically, the maximum absolute values of the partial derivatives over all channels).
This kind of saliency maps are obtained without effort as they only require a single backpropagation pass.
They also showed that this method is equivalent to the previously proposed deconvolutional nets method~\cite{zeiler2014vaucn} except for the difference on the calculation of ReLU layer's gradients.
Guided backpropagation~\cite{springenberg2014sfsta} combines above two methods.
It only takes into account the gradients (the former method) that have positive error signal (the latter method) when backpropagating through a ReLU layer.
There is also a variant Guided Grad-CAM (Gradient-weighted Class Activation Mapping)~\cite{selvaraju2017gcvef}, which first calculates a coarse-grained attribution map (with respect to a certain class) on the last convolutional layer and then multiply it to the attribution map obtained from guided backpropagation.
(Guided Grad-CAM is an extension of CAM\cite{zhou2016ldffd} which requires a special global average pooling layer.)

However, gradients themselves can be misleading.
Considering a piecewise continuous function,
\begin{equation*}
    y = \begin{cases}
        \bm{x}_1 + \bm{x}_2 & \text{if } \bm{x}_1 + \bm{x}_2 < 1 \,;    \\
        1                   & \text{if } \bm{x}_1 + \bm{x}_2 \geq 1 \,.
    \end{cases}
\end{equation*}
it is saturated when \(\bm{x}_1 + \bm{x}_2 \geq 1\).
At points where \(\bm{x}_1 + \bm{x}_2 > 1\), their gradients are always zeros.
DeepLIFT~\cite{shrikumar2017liftp} points out this problem and highlights the importance of having a reference input besides the target input to be explained.
The reference input is a kind of default or `neutral' input and will be different in different tasks (e.g., blank images or zero vectors).
Actually, as Sundararajan et al.~\cite{sundararajan2017aafdn} point out, DeepLIFT is trying to compute the ``discrete gradient'' instead of the (instantaneous) gradient.
Another similar ``discrete gradient'' method is LRP~\cite{bach2015opwef} (choosing a zero vector as the reference point), differing in how to compute the discrete gradient.
This view is also present in reference~\cite{ancona2018tbuog} that LRP and DeepLIFT are essentially computing backpropagation for modified gradient functions.

However, discrete gradients also have their drawbacks.
As the chain rule does not hold for discrete gradients, DeepLIFT and LRP adopt modified forms of backpropagation.
This makes their attributions specific to the network implementation, in other words, the attributions can be different even for two functionally equivalent networks (a concrete example can be seen in reference~\cite{sundararajan2017aafdn} appendix B).
Integrated gradients~\cite{sundararajan2017aafdn} have been proposed to address this problem.
It is defined as the path integral of all the gradients in the straight line between input \(\bm{x}\) and the reference input \(\bm{x}^{\text{ref}}\).
The \(i\)-th dimension of the integrated gradient (IG) is defined as follows,
\begin{equation*}
    \text{IG}_i(\bm{x}) \coloneqq \rbracket{\bm{x}_i - \bm{x}^{\text{ref}}_i} \int_0^1 \left.\frac{\partial f(\tilde{\bm{x}})}{\partial \tilde{\bm{x}}_{i}}\right|_{\tilde{\bm{x}}=\rbracket{\bm{x}^{\text{ref}}+\alpha\rbracket{\bm{x} - \bm{x}^{\text{ref}}}}} \,\mathrm{d}\alpha
\end{equation*}
where \(\frac{\partial f(\bm{x})}{\partial \bm{x}_i}\) is the \(i\)-th dimension of the gradient of \(f(\bm{x})\).
For those attribution methods requiring a reference point, semi-local interpretability is provided as users can select different reference points according to what to explain.
Table~\ref{tab:grad-attr} summarizes the above gradient-related attribution methods (adapted from~\cite{ancona2018tbuog}).
In addition to the ``gradient'' attribution discussed above, Wang et al.~\cite{wang2019bamba} point out that bias terms can contain attribution information complementary to the gradients.
They propose a method to recursively assign the bias attribution back to the input vector.

\begin{table}[!t]
    \renewcommand{\arraystretch}{1.3}
    \caption{Formulation of gradient-related attribution methods. {\upshape \(S_c\) is the output for class \(c\) (and it can be any neuron of interest), \(\sigma\) is the nonlinearity in the network and \(g\) is a replacement of \(\sigma'\) (the derivative of \(\sigma\)) in \(\frac{\partial S_c(\bm{x})}{\partial{\bm{x}}}\) in order to rewrite DeepLIFT and LRP with gradient formulation (see~\cite{ancona2018tbuog} for more details). \(\bm{x}_i\) is the \(i\)-th feature (pixel) of \(\bm{x}\).}}\label{tab:grad-attr}
    \centering
    \begin{tabularx}{\linewidth}{Xl}
        \toprule
        \tablehead{Method}                                  & \tablehead{Attribution}                                                                                                                                                                                                                 \\
        \midrule
        Gradient~\cite{baehrens2010hteic,simonyan2013dicnv} & \(\displaystyle \frac{\partial S_{c}(\bm{x})}{\partial \bm{x}_{i}}\)                                                                                                                                                                    \\[10pt]
        Gradient \(\odot\) Input                            & \(\displaystyle \bm{x}_{i} \cdot \frac{\partial S_{c}(\bm{x})}{\partial \bm{x}_{i}}\)                                                                                                                                                   \\[10pt]
        LRP~\cite{bach2015opwef}                            & \(\displaystyle \bm{x}_{i} \cdot \frac{\partial^{g} S_{c}(\bm{x})}{\partial \bm{x}_{i}},\;g=\frac{\sigma(\bm{z})}{\bm{z}}\)                                                                                                             \\[10pt]
        DeepLIFT~\cite{shrikumar2017liftp}                  & \(\displaystyle (\bm{x}_{i}-\bm{x}^{\text{ref}}_{i}) \cdot \frac{\partial^{g} S_{c}(\bm{x})}{\partial \bm{x}_{i}},\;g=\frac{\sigma(\bm{z})-\sigma(\bm{z}^{\text{ref}})}{\bm{z}-\bm{z}^{\text{ref}}}\)                                   \\[10pt]
        Integrated Gradient~\cite{sundararajan2017aafdn}    & \(\displaystyle (\bm{x}_{i}-\bm{x}^{\text{ref}}_{i}) \cdot \int_0^1 \left.\frac{\partial S_{c}(\tilde{\bm{x}})}{\partial{\tilde{\bm{x}}}_{i}}\right|_{\tilde{\bm{x}}=\bm{x}^{\text{ref}}+\alpha(\bm{x}-\bm{x}^{\text{ref}})} d \alpha\) \\
        \bottomrule
    \end{tabularx}
\end{table}

Those discrete gradient methods (e.g., LRP and DeepLIFT) provide semi-local explanations as they explain a target input w.r.t.\ another reference input.
But methods such as DeconvNet and Guided Backprop, which are only proposed to explain individual inputs, arguably have certain non-locality because of the rectification operation during the process.
Moreover, one can accumulate multiple local explanations to achieve a certain degree of global interpretability, which will be introduced in Section~\ref{sec:passive-attr-global}.

Although we have many approaches to produce plausible saliency maps, there is still a small gap between saliency maps and real explanations.
There have even been adversarial examples for attribution methods, which can produce perceptively indistinguishable inputs, leading to the \textit{same predicted labels}, but very \textit{different attribution maps}~\cite{ghorbani2019ionni,dombrowski2019ecbma,heo2019fnniv}.
Researchers came up with several properties a saliency map should have to be a valid explanation.
Sundararajan et al.~\cite{sundararajan2017aafdn} (integrated gradients method) introduced two requirements, \textit{sensitivity} and \textit{implementation invariance}.
Sensitivity requirement is proposed mainly because of the (local) gradient saturation problem (which results in zero gradient/attribution).
Implementation invariance means two functionally equivalent networks (which can have different learned parameters given the over-parametrizing setting of DNNs) should have the same attribution.
Kindermans et al.~\cite{kindermans2019tuosm} introduced input invariance.
It requires attribution methods to mirror the model's invariance with respect to transformations of the input.
For example, a model with a bias term can easily deal with a constant shift of the input (pixel values).
Obviously, (plain) gradient attribution methods satisfy this kind of input invariance.
For discrete gradient and other methods using reference points, they depend on the choices of reference.
Adebayo et al.~\cite{adebayo2018scfsm} took a different approach.
They found that edge detectors can also produce masks which look similar to saliency masks and highlight some features of the input.
But edge detectors have nothing to do with the network or training data.
Thus, they proposed two tests to verify whether the attribution method will fail (1) if the network's weights are replaced with random noise, or (2) if the labels of training data are shuffled.
The attribution methods should fail otherwise it suggests that the method does not reflect the trained network or the training data (in other words, it is just something like an edge detector).

\paragraph{Model agnostic attribution}

LIME~\cite{ribeiro2016wsity} is a well-known approach which can provide local attribution explanations (if choosing linear models as the so-called interpretable components).
Let \(f \colon \mathbb{R}^d \to \{+1,-1\}\) be a (binary classification) model to be explained.
Because the original input \(\bm{x} \in \mathbb{R}^d\) might be uninterpretable (e.g., a tensor of all the pixels in an image, or a word embedding~\cite{2013drowa}), LIME introduces an intermediate representation \(\bm{x}' \in \{0,1\}{}^{d'}\) (e.g., the existence of certain image patches or words).
\(\bm{x}'\) can be recovered to the original input space \(\mathbb{R}^d\).
For a given \(\bm{x}\), LIME tries to find a potentially interpretable model \(g\) (such as a linear model or decision tree) as a local explanation.
\begin{equation*}
    g_{\bm{x}} = \argmin_{g \in G} L(f,g,\pi_{\bm{x}}) + \Omega(g)
\end{equation*}
where \(G\) is the explanation model family, \(L\) is the loss function that measures the fidelity between \(f\) and \(g\).
\(L\) is evaluated on a set of perturbed samples around \(\bm{x}'\) (and their recovered input), which are weighted by a local kernel \(\pi_{\bm{x}}\).
\(\Omega{}\) is the complexity penalty of \(g\), ensuring \(g\) to be interpretable.
%% MAPLE
MAPLE~\cite{plumb2018masle} is a similar method using local linear models as explanations.
The difference is it defines the locality as how frequently the data points fall into a same leaf node in a proxy random forest (fit on the trained network).

In game theory, there is a task to ``fairly'' assign each player a payoff from the total gain generated by a coalition of all players.
Formally, let \(N\) be a set of \(n\) players, \(v \colon 2^N \to \mathbb{R}\) is a characteristic function, which can be interpreted as the total gain of the coalition \(N\).
Obviously, \(v(\varnothing) = 0\).
A coalitional game can be denoted by the tuple \(\langle N,v \rangle{}\).
Given a coalitional game, Shapley value~\cite{shapley1953avfnp} is a solution to the payoff assignment problem.
The payoff (attribution) for player \(i\) can be computed as follows,
\begin{equation*}
    \phi_{i}(v)=\frac{1}{|N|} \sum_{S \subseteq N\backslash \{i\}}\binom{|N|-1}{|S|}^{-1}(v(S \cup \{i\}) - v(S))
\end{equation*}
where \(v(S \cup \{i\}) - v(S)\) is the marginal contribution of player \(i\) to coalition \(S\).
The rest of the formula can be viewed as a normalization factor.
A well-known alternative form of Shapley value is
\begin{equation*}
    \phi_{i}(v)=\frac{1}{|N| !} \sum_{O \in \mathfrak{S}(N)}\sbracket{v\rbracket{P_i^O \cup{}\{i\}} - v\rbracket{P_i^O}}
\end{equation*}
where \(\mathfrak{S}(N)\) is the set of all ordered permutations of \(N\), and \(P_{i}^{O}\) is the set of players in \(N\) which are predecessors of player \(i\) in the permutation \(O\).
\v{S}trumbelj and Kononenko adopted this form so that \(v\) can be approximated in polynomial time~\cite{strumbelj2010aeeoi} (also see~\cite{ancona2019ednnw} for another approximation method).

Back to the neural network (denoted by \(f\)), let \(N\) be all the input features (attributes), \(S\) is an arbitrary feature subset of interest (\(S \subseteq N\)).
For an input \(\bm{x}\), the characteristic function \(v(S)\) is the difference between the expected model output when we know all the features in \(S\), and the expected output when no feature value is known (i.e.\ the expected output over all possible input), denoted by
\begin{equation*}
    v(S) = \frac{1}{|\mathcal{X}^{N \setminus S}|}\sum_{\bm{y} \in \mathcal{X}^{N \setminus S}}f(\tau(\bm{x},\bm{y},S)) - \frac{1}{|\mathcal{X}^N|}\sum_{\bm{z} \in \mathcal{X}^N}f(\bm{z})
\end{equation*}
\(\mathcal{X}^N\) and \(\mathcal{X}^{N \setminus S}\) are respectively the input space containing feature sets \(N\) and \(N \setminus S\).
\(\tau(\bm{x},\bm{y},S)\) is a vector composed by \(\bm{x}\) and \(\bm{y}\) according to whether the feature is in \(S\).

However, a practical problem is the exponential computation complexity, let alone the cost of the feed-forward computing on each \(v(S)\) call.
\v{S}trumbelj and Kononenko~\cite{strumbelj2010aeeoi} approximate Shapley value by sampling from \(\mathfrak{S}(N)\times\mathcal{X}\) (Cartesian product).
There are other variants such as using different \(v\).
More can be found in reference~\cite{lundberg2017auati} which proposes a unified view including not only the Shapley value methods but also LRP and DeepLIFT\@.
There is also Shapley value through the lens of causal graph~\cite{heskes2020csvec}.

Sensitivity analysis can also be used to evaluate the importance of a feature.
Specifically, the importance of a feature could be measured as how much the model output will change upon the change of a feature (or features).
There are different kinds of changes, e.g., perturbation, occlusion~\cite{zeiler2014vaucn,fong2017ieobb} etc.~\cite{zintgraf2017vdnnd}.

Chen et al.~\cite{chen2018lteai} propose an instance-wise feature selector \(\mathcal{E}\) which maps an input \(\bm{x}\) to a conditional distribution \(P(S \mid \bm{x})\), where \(S\) is any subset (of certain size) of the original feature set.
The selected features can be denoted by \(\bm{x}_S\).
Then they aim to maximize the mutual information between the selected features and the output variable \(Y\),
\[\max_\mathcal{E} I(X_S;Y) \quad \text{subject to} \quad S \sim \mathcal{E}(X).\]
A variational approximation is used to obtain a tractable solution of the above problem.

\subsubsection{Passive, Attribution as Explanation, \textbf{Global}}\label{sec:passive-attr-global}

A natural way to get global attribution is to combine individual ones obtained from above local/semi-local methods.
SpRAy~\cite{lapuschkin2019uchpa} clusters on the individual attributions and then summarizes some groups of prediction strategies.
MAME~\cite{natesan2020mame} is a similar method that can generate a multilevel (local to global) explanation tree.
Salman et al.~\cite{salman2020deep} provide a different way, which makes use of multiple neural networks.
Each of the network can provide its own local attributions, on top of which a clustering is performed.
Those clusters, intuitively the consensus of multiple models, can provide more robust interpretations.

The attribution does not necessarily attribute `credits' or `blame' to the \textit{raw} input or features.
Kim et al.~\cite{kim2018ibfaq} propose a method TCAV (quantitative Testing with Concept Activation Vectors) that can compute the model sensitivity of any user-interested concept.
By first collecting some examples with and without a target concept (e.g., the presence of stripes in an animal), the concept can then be represented by a normal vector to the hyperplane separating those positive/negative examples (pictures of animals with/without stripes) in a hidden layer.
The score of the concept can be computed as the (average) output sensitivity if the hidden layer representation (of an input \(\bm{x}\)) moves an infinitesimally small step along the concept vector.
This is a global interpretability method as it explains how a concept affects the output in general.
Besides being manually picked by a human, these concepts can also be discovered automatically by clustering input segments~\cite{ghorbani2019tacbe}.

\subsection{Passive, \textbf{Explanation by Example}}

The last kind of explanations we reviewed is explanation by example.
When asked for an explanation for a new input, these approaches return other example(s) for supporting or counter example.
One basic intuition is to find examples that the model considers to be most similar (in terms of latent representations)~\cite{caruana1999cbeon}.
This is \textit{local} interpretability but we can also seek a set of representative samples within a class or for more classes that provides \textit{global} interpretability.
A general approach is presented in~\cite{2011psfic}.
{There are other methods, such as measuring how much a training example affects the model prediction on a target input.}
Here we only focus on work related to deep neural networks.

\subsubsection{Passive, Explanation by Example, \textbf{Local}}

Koh and Liang~\cite{koh2017ubbpv} provide an interesting method to evaluate how much a training example affects the model prediction on an unseen test example.
The change of model parameters upon a change of training example is first calculated with approximation.
Further, its influence on the loss at the test point can be computed.
By checking the most influential (positively or negatively) training examples (for the test example), we can have some insights on the model predictions.
Yeh et al.~\cite{yeh2018rpsfe} show that the logit (the neuron before softmax) can be decomposed into a linear combination of training points' activations in the pre-logit layer.
The coefficients of the training points indicate whether the similarity to those points is excitatory or inhibitory.
The above two approaches both provide local explanations.

\section{Active Interpretability Intervention During Training}\label{sec:active}

Besides passive looking for human-understandable patterns from the trained network, researchers also tried to impose interpretability restrictions during the network training process, i.e.\ active interpretation methods in our taxonomy.
A popular idea is to add a special regularization term \(\Omega(\theta)\) to the loss function, also known as ``interpretability loss'' (\( \theta \) collects all the weights of a network).
We now discuss the related papers according to the forms of explanations they provide.

\subsection{Active, \textbf{Rule as Explanation} (semi-local or global)}

Wu et al.~\cite{wu2018bstro} propose tree regularization which favours models that can be well approximated by shallow decision trees.
It requires two steps
(1) train a binary decision tree using data points \(\cbracket{\bm{x}^{(i)}, \hat{y}^{(i)}}^N\), where \(\hat{y} = f_{\theta}(\bm{x})\) is the network prediction rather than the true label,
(2) calculate the average path length (from root to leaf node) of this decision tree over all the data points.
However, this tree regularization term \(\Omega(\theta)\) is not differentiable.
Therefore, a surrogate regularization term \(\hat{\Omega}(\theta)\) was introduced.
Given a dataset \(\cbracket{\theta^{(j)}, \Omega(\theta^{(j)})}^J_{j=1}\), \(\hat{\Omega}\) can be trained as a multi-layer perceptron network which minimizes the squared error loss
\begin{equation*}
    \min _{\xi} \sum_{j=1}^{J}\rbracket{\Omega\rbracket{\theta^{(j)}} - \hat{\Omega}\rbracket{\theta^{(j)}; \xi}}^{2}+\epsilon\left\|\xi\right\|_{2}^{2}
\end{equation*}
\(\cbracket{\theta^{(j)}, \Omega(\theta^{(j)})}^J_{j=1}\) can be assembled during network training.
Also, data augmentation techniques can be used to generate \( \theta \), especially in the early training phase.
Tree regularization enables global interpretability as it forces a network to be easily approximable by a decision tree.
Later, the authors also proposed regional tree regularization which did this in a semi-local way~\cite{wu2020rtrfi}.

\subsection{Active, \textbf{Hidden semantics as Explanation} (global)}

Another method aims to make a convolutional neural network learn better (disentangled) hidden semantics.
Having seen feature visualization techniques mentioned above and some empirical studies~\cite{bau2017ndqio,zhou2018idvrv}, CNNs are believed to have learned some low-level to high-level representations in the hierarchical structure.
But even if higher-layers have learned some object-level concepts (e.g., head, foot), those concepts are usually entangled with each other.
In other words, a high-layer filter may contain a mixture of different patterns.
Zhang et al.~\cite{zhang2018icnn} propose a loss term which encourages high-layer filters to represent a single concept.
Specifically, for a CNN, a feature map (output of a high-layer filter, after ReLU) is an \(n \times n\) matrix.
Zhang et al.\ predefined a set of \(n^2\) ideal feature map templates (activation patterns) \(\mathbf{T}\), each of which is like a Gaussian kernel only differing on the position of its peak.
During the forward propagation, the feature map is masked (element-wise product) by a certain template \(\mathrm{T} \in \mathbf{T}\) according to the position of the most activated ``pixel'' in the original feature map.
During the back propagation, an extra loss is plugged in, which is the mutual information between \(\mathbf{M}\) (the feature maps of a filter calculated on all images) and \( \mathbf{T} \cup \{\mathrm{T^-}\} \) (all the ideal activation patterns plus a negative pattern which is full of a negative constant).
This loss term makes a filter to either have a consistent activation pattern or keep inactivated.
Experiments show that filters in their designed architecture are more semantically meaningful (e.g., the ``receptive field''~\cite{zhou2015odeid} of a filter corresponds to the head of animals).

\subsection{Active, \textbf{Attribution as Explanation}}

Similar to tree regularization which helps to achieve better global interpretability (decision trees), ExpO~\cite{plumb2020rbbmf} added an interpretability regularizer in order to improve the quality of local attribution.
That regularization requires a model to have fidelitous (high fidelity) and stable local attribution.
DAPr~\cite{weinberger2020ldapb} (deep attribution prior) took into account additional information (e.g., a rough prior about the feature importance).
The prior will be trained jointly with the main prediction model (as a regularizer) and biases the model to have similar attribution as the prior.

Besides performing attribution on individual input (\textit{locally} in input space), Dual-net~\cite{wojtas2020firfd} was proposed to decide feature importance population-wise, i.e., finding an `optimal' feature subset collectively for an input population.
In this method, a \textit{selector} network is used to generate an optimal feature subset, while an \textit{operator} network makes predictions based on that feature set.
These two networks are trained jointly.
After training, the selector network can be used to rank feature importance.

\subsection{Active, \textbf{Explanations by Prototypes} (global)}

Li et al.~\cite{li2018dlfcb} incorporated a prototype layer to a network (specifically, an autoencoder).
The network acts like a prototype classifier, where predictions are made according to the proximity between (the encoded) inputs and the learned prototypes.
Besides the cross-entropy loss and the (autoencoder) reconstruction error, they also included two interpretability regularization terms, encouraging every prototype to be similar to at least one encoded input, vice versa.
After the network is trained, those prototypes can be naturally used as explanations.
Chen et al.~\cite{chen2019tlltd} add a prototype layer to a regular CNN rather than an autoencoder.
This prototype layer contains prototypes that are encouraged to resemble parts of an input.
When asked for explanations, the network can provide several prototypes for different parts of the input image respectively.

\section{Evaluation of Interpretability}\label{sec:eval}

In general, interpretability is hard to evaluate objectively as the end-tasks can be quite divergent and may require domain knowledge from experts~\cite{montavon2018mfiau}.
Doshi-Velez and Kim~\cite{doshi2017tarso} proposed three evaluation approaches: application-grounded, human-grounded, and functionally-grounded.
The first one measures to what extent interpretability helps the end-task (e.g., better identification of errors or less discrimination).
Human-grounded approaches are, for example, directly letting people evaluate the quality of explanations with human-subject experiments (e.g., let a user choose which explanation is of the highest quality among several explanations).
Functionally-grounded methods find proxies for the explanation quality (e.g., sparsity).
The last kind of approaches require no costly human experiments but how to properly determine the proxy is a challenge.

In our taxonomy, explanations are divided into different types.
Although the interpretability can hardly be compared between different types of explanations, there are some measurements proposed for this purpose.
For logic rules and decision trees, the size of the extracted rule model is often used as a criterion~\cite{tickle1998ttwct,odajima2008grgfd,wu2018bstro} (e.g., the number of rules, the number of antecedents per rule, the depth of the decision tree etc.).
Strictly speaking, these criteria measure more about whether the explanations are efficiently interpretable.
Hidden semantics approaches produce explanations on certain hidden units in the network.
Network Dissection~\cite{bau2017ndqio} quantifies the interpretability of hidden units by calculating their matchiness to certain concepts.
As for the hidden unit visualization approaches, there is no a good measurement yet.
For attribution approaches, their explanations are saliency maps/masks (or feature importance etc.\ according to the specific task).
Samek et al.~\cite{samek2016etvow} evaluate saliency maps by the performance degradation if the input image is partially masked with noise in an order from salient to not salient patches.
A similar evaluation method is proposed in~\cite{adebayo2018scfsm} and Hooker et al.~\cite{hooker2019abfim} suggest using a fixed uninformative value rather than noise as the mask and evaluating performance degradation on a retrained model.
Samek et al.\ also use entropy as another measure in the belief that good saliency maps focus on relevant regions and do not contain much irrelevant information and noise.
Montavon et al.~\cite{montavon2018mfiau} would like the explanation function (which maps an input to a saliency map) to be continuous/smooth, which means the explanations (saliency maps) should not vary too much when seeing similar inputs.

\section{Discussion}

In practice, different interpretation methods have their own advantages and disadvantages.
Passive (post-hoc) methods have been widely studied because they can be applied in a relatively straightforward manner to most existing networks.
One can choose methods that make use of a network's inner information (such as connection weights, gradients), which are usually more efficient (e.g., see Paragraph~\ref{par:gradient-attribution}).
Otherwise there are also model-agnostic methods that have no requirement of the model architecture, which usually compute the marginal effect of a certain input feature.
But this generality is also a downside of passive methods, especially because there is no easy way to incorporate specific domain knowledge/priors.
Active (interpretability intervention) methods put forward ideas about how a network should be optimized to gain interpretability.
The network can be optimized to be easily fit by a decision tree, or to have preferred feature attribution, better tailored to a target task.
However, the other side of the coin is that such active intervention requires the compatibility between networks and interpretation methods.

As for the second dimension, the format of explanations, logical rules are the most clear (do not need further human interpretation).
However, one should carefully control the complexity of the explanations (e.g., the depth of a decision tree), otherwise the explanations will not be useful in practice.
Hidden semantics essentially explain a subpart of a network, with most work developed in the computer vision field.
Attribution is very suitable for explaining individual inputs.
But it is usually hard to get some overall understanding about the network from attribution (compared to, e.g., logical rules).
Explaining by providing an example has the lowest (the most implicit) explanatory power.

As for the last dimension, local explanations are more useful when we care more about every single prediction (e.g., a credit or insurance risk assessment).
For some research fields, such as genomics and astronomy, global explanations are more preferred as they may reveal some general ``knowledge''.
Note, however, there is no hard line separating local and global interpretability.
With the help of explanation fusing methods (e.g., MAME), we can obtain multilevel (from local to global) explanations.

\section{Conclusion}\label{sec:conclusion}

In this survey, we have provided a comprehensive review of neural network interpretability.
First, we have discussed the definition of interpretability and stressed the importance of the format of explanations and domain knowledge/representations.
Specifically, there are four commonly seen types of explanations: \textit{logic rules}, \textit{hidden semantics}, \textit{attribution} and \textit{explanations by examples}.
Then, by reviewing the previous literature, we summarized 3 essential reasons why interpretability is important: the requirement of high reliability systems, ethical/legal requirements and knowledge finding for science.
After that, we introduced a novel taxonomy for the existing network interpretation methods.
It evolves along three dimensions: passive vs.\ active, types of explanations and global vs.\ local interpretability.
The last two dimensions are not purely categorical but with ordinal values (e.g., semi-local).
This is the first time we have a coherent overview of interpretability research rather than many isolated problems and approaches.
We can even visualize the distribution of the existing approaches in the 3D space spanned by our taxonomy.

From the perspective of the new taxonomy, there are still several possible research directions in the interpretability research.
First, the active interpretability intervention approaches are underexplored.
Some analysis of the passive methods also suggests that the neural network does not necessarily learn representations which can be easily interpreted by human beings.
Therefore, how to actively make a network interpretable without harming its performance is still an open problem.
During the survey process, we have seen more and more recent work filling this blank.

Another important research direction may be how to better incorporate domain knowledge in the networks.
As we have seen in this paper, interpretability is about providing explanations.
And explanations build on top of understandable terms (or concepts) which can be specific to the targeted tasks.
We already have many approaches to construct explanations of different types, but the domain-related terms used in the explanations are still very simple (see Table~\ref{tab:terms}).
If we can make use of terms that are more domain/task-related, we can get more informative explanations and better interpretability.

% if have a single appendix:
%\appendix[Proof of the Zonklar Equations]
% or
%\appendix  % for no appendix heading
% do not use \section anymore after \appendix, only \section*
% is possibly needed

% use appendices with more than one appendix
% then use \section to start each appendix
% you must declare a \section before using any
% \subsection or using \label (\appendices by itself
% starts a section numbered zero.)
%

% \appendices
% \section{Proof of the First Zonklar Equation}
% Appendix one text goes here.

% % you can choose not to have a title for an appendix
% % if you want by leaving the argument blank
% \section{}
% Appendix two text goes here.

% use section* for acknowledgment
% \section*{Acknowledgment}

% Can use something like this to put references on a page
% by themselves when using endfloat and the captionsoff option.
\ifCLASSOPTIONcaptionsoff
  \newpage
\fi

% trigger a \newpage just before the given reference
% number - used to balance the columns on the last page
% adjust value as needed - may need to be readjusted if
% the document is modified later
%\IEEEtriggeratref{8}
% The "triggered" command can be changed if desired:
%\IEEEtriggercmd{\enlargethispage{-5in}}

% references section

% can use a bibliography generated by BibTeX as a .bbl file
% BibTeX documentation can be easily obtained at:
% http://mirror.ctan.org/biblio/bibtex/contrib/doc/
% The IEEEtran BibTeX style support page is at:
% http://www.michaelshell.org/tex/ieeetran/bibtex/
\bibliographystyle{IEEEtran}
% argument is your BibTeX string definitions and bibliography database(s)
\bibliography{IEEEabrv,ref}

% biography section
% 
% If you have an EPS/PDF photo (graphicx package needed) extra braces are
% needed around the contents of the optional argument to biography to prevent
% the LaTeX parser from getting confused when it sees the complicated
% \includegraphics command within an optional argument. (You could create
% your own custom macro containing the \includegraphics command to make things
% simpler here.)
%\begin{IEEEbiography}[{\includegraphics[width=1in,height=1.25in,clip,keepaspectratio]{mshell}}]{Michael Shell}
% or if you just want to reserve a space for a photo:

%% Last page column equalisation
% \enlargethispage{-3in}

\begin{IEEEbiography}[{\includegraphics[width=1in,height=1.25in,clip,keepaspectratio]{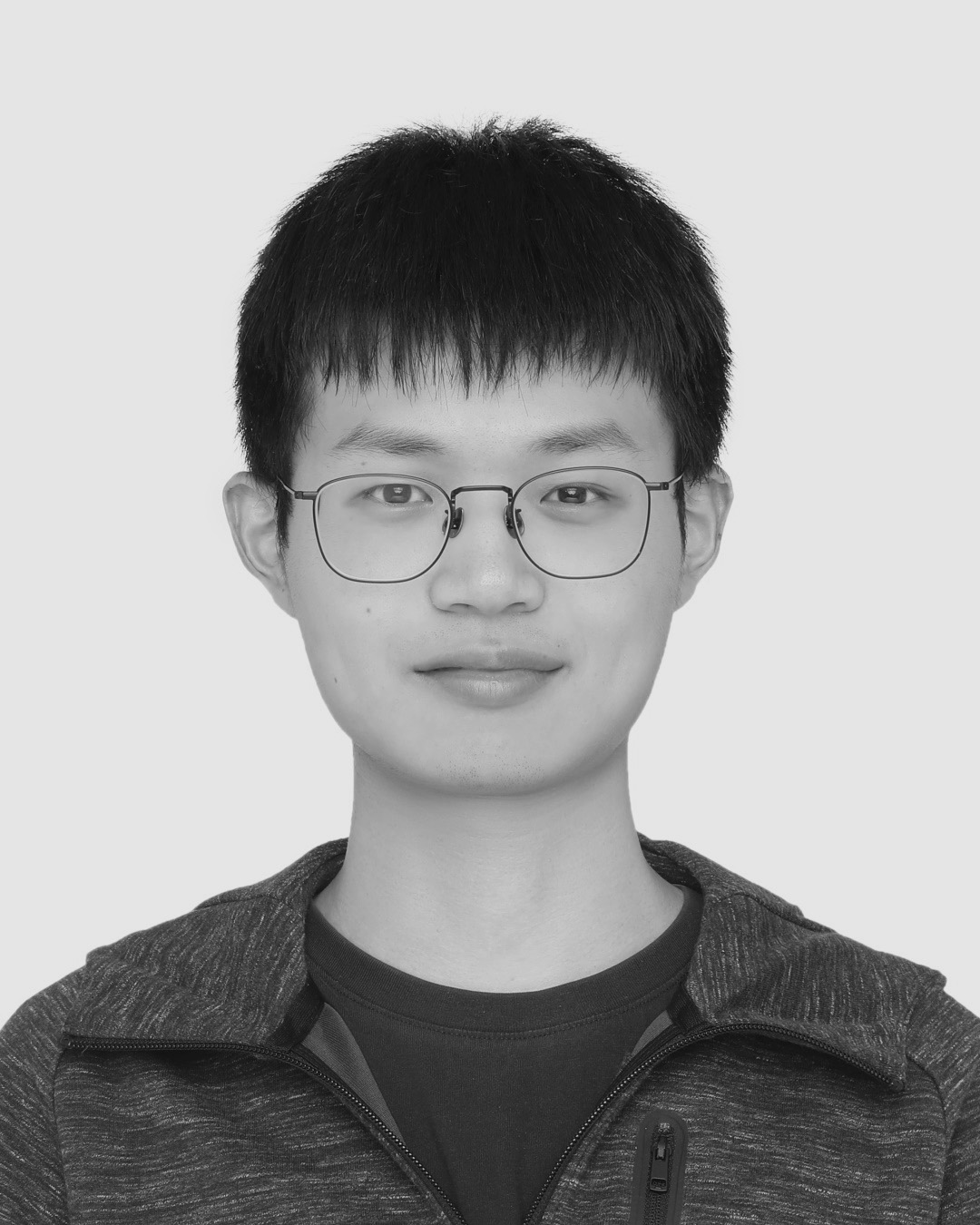}}]{Yu Zhang}
received the B.Eng. degree from the Department of Computer Science and Engineering,
Southern University of Science and Technology, China, in 2017.
He is currently pursuing the Ph.D. degree in the Department of Computer Science and Engineering, Southern University of Science and Technology, jointly with the School of Computer Science, University of Birmingham,
Edgbaston, Birmingham, UK\@.
His current research interest is interpretable machine learning.
\end{IEEEbiography}

\begin{IEEEbiography}[{\includegraphics[width=1in,height=1.25in,clip,keepaspectratio]{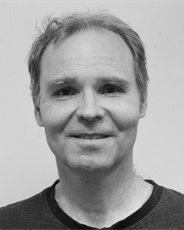}}]{Peter Tino}
(M.Sc.\ Slovak University of Technology, Ph.D.\ Slovak Academy of Sciences) was a Fulbright Fellow with the NEC Research Institute,
Princeton, NJ, USA, and a Post-Doctoral Fellow with the Austrian Research Institute for AI, Vienna, Austria, and with Aston University, Birmingham, U.K.
Since 2003, he has been with the School of Computer Science, University of Birmingham, Edgbaston, Birmingham, U.K., where he is currently a full Professor---Chair in Complex and Adaptive Systems.
His current research interests include dynamical systems, machine learning, probabilistic modelling of structured data, evolutionary computation, and fractal analysis.
Peter was a recipient of the Fulbright Fellowship in 1994, the U.K.–Hong-Kong Fellowship for Excellence in 2008, three Outstanding Paper of the Year Awards from the IEEE Transactions on Neural Networks in 1998 and 2011 and the IEEE Transactions on Evolutionary Computation in 2010, and the Best Paper Award at ICANN 2002.
He serves on the editorial boards of several journals.
\end{IEEEbiography}

\begin{IEEEbiography}[{\includegraphics[width=1in,height=1.25in,clip,keepaspectratio]{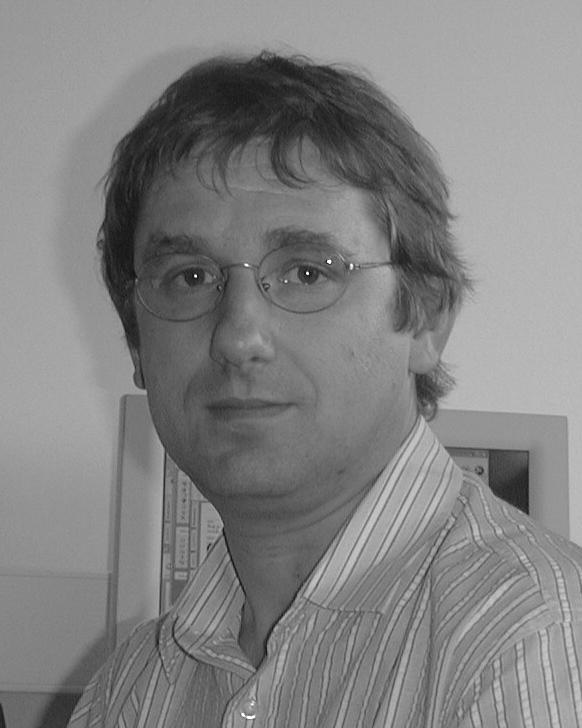}}]{Ale\v{s} Leonardis}
is Chair of Robotics at the School of Computer Science, University of Birmingham.
He is also Professor of Computer and Information Science at the University of Ljubljana.
He was a visiting researcher at the GRASP Laboratory at the University of Pennsylvania, post-doctoral fellow at PRIP Laboratory, Vienna University of Technology, and visiting professor at ETH Zurich and University of Erlangen.
His research interests include computer vision, visual learning, and biologically motivated vision---all in a broader context of cognitive systems and robotics.
Ale\v{s} Leonardis was a Program Co-chair of the European Conference on Computer Vision 2006, and he has been an Associate Editor of the IEEE PAMI and IEEE Robotics and Automation Letters, an editorial board member of Pattern Recognition and Image and Vision Computing, and an editor of the Springer book series Computational Imaging and Vision.
In 2002, he coauthored a paper, Multiple Eigenspaces, which won the 29th Annual Pattern Recognition Society award.
He is a fellow of the IAPR and in 2004 he was awarded one of the two most prestigious national (SI) awards for his research achievements.
\end{IEEEbiography}

\begin{IEEEbiography}[{\includegraphics[width=1in,height=1.25in,clip,keepaspectratio]{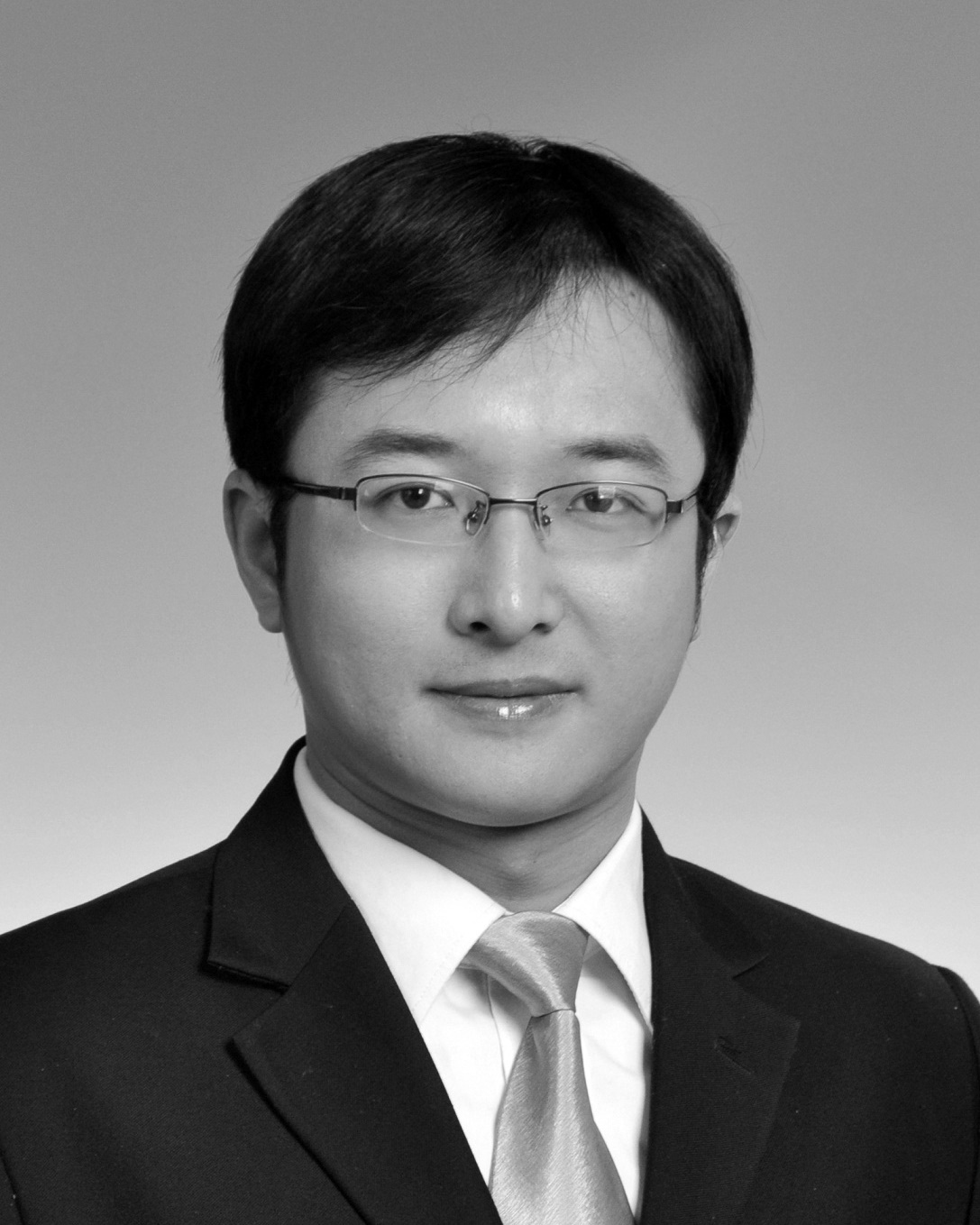}}]{Ke Tang}
(Senior Member, IEEE) received the B.Eng.\ degree from the Huazhong University of Science and Technology, Wuhan, China, in 2002 and the Ph.D.\ degree from Nanyang Technological University, Singapore, in 2007.
From 2007 to 2017, he was with the School of Computer Science and Technology, University of Science and Technology of China, Hefei, China, first as an Associate Professor from 2007 to 2011 and later as a Professor from 2011 to 2017.
He is currently a Professor with the Department of Computer Science and Engineering, Southern University of Science and Technology, Shenzhen, China.
He has over 10000 Google Scholar citation with an H-index of 48.
He has published over 70 journal papers and over 80 conference papers.
His current research interests include evolutionary computation, machine learning, and their applications.
Dr.\ Tang was a recipient of the Royal Society Newton Advanced Fellowship in 2015 and the 2018 IEEE Computational Intelligence Society Outstanding Early Career Award.
He is an Associate Editor of the IEEE Transactions on Evolutionary Computation and served as a member of Editorial Boards for a few other journals.
\end{IEEEbiography}

% insert where needed to balance the two columns on the last page with
% biographies
%\newpage

% You can push biographies down or up by placing
% a \vfill before or after them. The appropriate
% use of \vfill depends on what kind of text is
% on the last page and whether or not the columns
% are being equalized.

% \vfill

% Can be used to pull up biographies so that the bottom of the last one
% is flush with the other column.
%\enlargethispage{-5in}

\end{document}

%% file: imgs/taxonomy.tikz
\begin{tikzpicture}[every node/.style={anchor=west},thick]
    \definecolor{C0}{HTML}{1F77B4}
    \definecolor{C1}{HTML}{FF7F0E}
    \definecolor{C2}{HTML}{2CA02C}
    \def\h{\baselineskip}
    \fill[C0!20](-0.3,0.8\h) rectangle (15.5,-3.6\h);
    \fill[C0!40](-0.3,0.8\h) rectangle (15.5,-0.6\h);
    \draw(-0.1,0) node[very thick,inner sep=0,outer sep=0] {\strut \textbf{Dimension 1 --- Passive vs.\ Active Approaches}};
    \draw[decorate,decoration=brace](1.25pt,-3.25\h) -- (1.25pt,-0.95\h);
    \draw(0.1,-1.5\h) node {\strut Passive};
    \draw(0.1,-2.8\h) node {\strut Active};
    \draw(3,-1.5\h) node[node font=\small,color=black!75] {\strut Post hoc explain trained neural networks};
    \draw(3,-2.8\h) node[node font=\small,color=black!75] {\strut Actively change the network architecture or training process for better interpretability};
    \fill[C1!20](-0.3,-3.9\h) rectangle (15.5,-12\h);
    \fill[C1!40](-0.3,-3.9\h) rectangle (15.5,-5.3\h);
    \node (dim2) [very thick,inner sep=0,outer sep=0] at (-0.1,-4.7\h) {\strut \textbf{Dimension 2 --- Type of Explanations}};
    \draw ($(dim2.east) + (0.1,0)$) node [inner sep=0,outer sep=0] {(in the order of increasing explanatory power)};
    \draw(-0.1,-6.1\h) node[inner sep=0,outer sep=0] {\strut To explain a prediction/class by};
    \draw[-latex](0,-6.9\h) -- (0,-11.7\h);
    \draw(-0.1,-6.9\h) -- +(0.2,0);
    \draw(0.1,-7.4\h) node {\strut Examples};
    \draw(0.1,-8.7\h) node {\strut Attribution};
    \draw(0.1,-10\h) node {\strut Hidden semantics};
    \draw(0.1,-11.3\h) node {\strut Rules};
    \draw(3,-7.4\h) node[node font=\small,color=black!75] {\strut Provide example(s) which may be considered similar or as prototype(s)};
    \draw(3,-8.7\h) node[node font=\small,color=black!75] {\strut Assign credit (or blame) to the input features (e.g.\ feature importance, saliency masks)};
    \draw(3,-10\h) node[node font=\small,color=black!75] {\strut Make sense of certain hidden neurons/layers};
    \draw(3,-11.3\h) node[node font=\small,color=black!75] {\strut Extract logic rules (e.g.\ decision trees, rule sets and other rule formats)};
    \fill[C2!20](-0.3,-12.3\h) rectangle (15.5,-17.8\h);
    \fill[C2!40](-0.3,-12.3\h) rectangle (15.5,-13.7\h);
    \node (dim3) [very thick,inner sep=0,outer sep=0] at (-0.1,-13.1\h) {\strut \textbf{Dimension 3 --- Local vs.\ Global Interpretability}};
    \draw ($(dim3.east) + (0.1,0)$) node [inner sep=0,outer sep=0] {(in terms of the input space)};
    \draw[-latex](0,-14\h) -- (0,-17.5\h);
    \draw(-0.1,-14\h) -- +(0.2,0);
    \draw(0.1,-14.5\h) node {\strut Local};
    \draw(0.1,-15.8\h) node {\strut Semi-local};
    \draw(0.1,-17.1\h) node {\strut Global};
    \draw(3,-14.5\h) node[node font=\small,color=black!75] {\strut Explain network's \textit{predictions on individual samples} (e.g.\ a saliency mask for an input image)};
    \draw(3,-15.8\h) node[node font=\small,color=black!75] {\strut In between, for example, explain a group of similar inputs together};
    \draw(3,-17.1\h) node[node font=\small,color=black!75] {\strut Explain the network \textit{as a whole} (e.g.\ a set of rules/a decision tree)};
\end{tikzpicture}

%% file: imgs/cell22.tikz
\begin{tikzpicture}[every node/.style={inner sep=0,outer sep=0},thick]
    \definecolor{C0}{HTML}{1F77B4}
    \def\h{\baselineskip}
    \draw (0,0) node [anchor=west] {\textbullet{} An example active method~\cite{zhang2018icnn} adds a special loss term};
    \draw (0,-1\h) node [anchor=west] {that encourages filters to learn consistent and exclusive};
    \draw (0,-2\h) node [anchor=west] {patterns (e.g.\ head patterns of animals)};
    \node (a) at (1.35,-3.4\h) {};
    %% connections
    \foreach \x in {0,1,2,3}
        \foreach \y in {0,1,2}
            \draw [black!60] ($(a)+(0,-0.4*\x)$) -- ($(a)+(0.5,-0.2-0.4*\y)$);
    \foreach \x in {0,1,2}
        \foreach \y in {0,1,2}
            \draw [black!60] ($(a)+(0.5,-0.2-0.4*\x)$) -- ($(a)+(1,-0.2-0.4*\y)$);
    \foreach \x in {0,1,2}
        \foreach \y in {0,1}
            \draw [black!60] ($(a)+(1,-0.2-0.4*\x)$) -- ($(a)+(1.5,-0.4-0.4*\y)$);
    %% neurons
    \foreach \x in {0,1,2,3}
        \draw [black!60,fill=white] ($(a)+(0,-0.4*\x)$) circle (0.15);
    \foreach \x in {0,1,2}
        \draw [black!60,fill=white] ($(a)+(0.5,-0.2-0.4*\x)$) circle (0.15);
    \foreach \x in {0,1}
        \draw [black!60,fill=white] ($(a)+(1,-0.2-0.4*\x)$) circle (0.15);
    \node (neuron) at ($(a)+(1,-0.2-0.4-0.4)$) {};
    \draw [C0,fill=white] (neuron) circle (0.15);
    \foreach \x in {0,1}
        \draw [black!60,fill=white] ($(a)+(1.5,-0.4-0.4*\x)$) circle (0.15);
    %% input output
    \node [black!60,anchor=east,align=right] (in) at ($(a)+(-0.15,-0.6)+(-0.4,0)$) {\strut image};
    \draw [black!60,-latex] ($(in.east)+(0.05,0)$) -- +(0.3,0);
    \node [black!60,anchor=west,align=left] (in) at ($(a)+(1.65,-0.6)+(0.4,0)$) {\strut animal label};
    \draw [black!60,-latex] ($(a)+(1.65,-0.6)+(0.05,0)$) -- +(0.3,0);
    \node [align=center,C0] (text) at ($(neuron)+(0,-0.6)$) {actual\\``receptive fields''~\cite{zhou2015odeid}:};
    \draw ($(text.east)+(0.15,0)$) node [anchor=west] {\includegraphics[width=0.8cm]{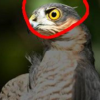}};
    \draw ($(text.east)+(1.1,0)$) node [anchor=west] {\includegraphics[width=0.8cm]{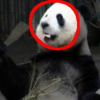}};
    \draw ($(text.east)+(2.05,0)$) node [anchor=west] {\includegraphics[width=0.8cm]{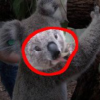}};
\end{tikzpicture}

%% file: imgs/cell31.tikz
\begin{tikzpicture}[every node/.style={inner sep=0,outer sep=0,anchor=west},thick]
    \definecolor{C0}{HTML}{1F77B4}
    \def\h{\baselineskip}
    \node (a) at (0,0) {For \(\bm{x}^{(i)}\):};
    \node (b) at ($(a.east)+(0.05,0)$) {\includegraphics[width=0.8cm]{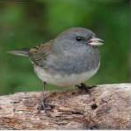}};
    \draw [-latex] ($(b.east)+(0.05,0)$) -- +(0.3,0);
    \node [inner sep=5pt,draw] (nn) at ($(b.east)+(0.4,0)$) {neural net};
    \draw [-latex] ($(nn.east)+(0.05,0)$) -- +(0.3,0);
    \node (c) at ($(nn.east)+(0.4,0)$) {\(\hat{y}^{(i)}\): junco bird};
    \node (a1) at (0,-0.8) {The ``contribution''\textsuperscript{1} of each pixel:};
    \draw ($(a1.east)+(0.08,0)$) node [anchor=west] {\includegraphics[width=0.8cm]{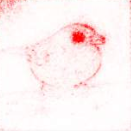}};
    \draw ($(a1.east)+(0.93,0)$) node [anchor=west] {\cite{adebayo2018scfsm}};
    \node (a2) at (0,-1.45) {a.k.a.\ saliency map, which can be computed by different};
    \draw ($(a2.west)+(0,-\h)$) node {methods like gradients~\cite{simonyan2013dicnv}, sensitivity analysis\textsuperscript{2}~\cite{zeiler2014vaucn} etc.};
\end{tikzpicture}

%% file: imgs/cell41.tikz
\begin{tikzpicture}[every node/.style={inner sep=0,outer sep=0,anchor=west},thick]
    \definecolor{C0}{HTML}{1F77B4}
    \def\h{\baselineskip}
    \node (a) at (0,0) {For \(\bm{x}^{(i)}\):};
    \node (b) at ($(a.east)+(0.05,0)$) {\includegraphics[width=0.8cm]{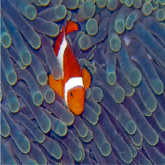}};
    \draw [-latex] ($(b.east)+(0.05,0)$) -- +(0.3,0);
    \node [inner sep=5pt,draw] (nn) at ($(b.east)+(0.4,0)$) {neural net};
    \draw [-latex] ($(nn.east)+(0.05,0)$) -- +(0.3,0);
    \node (c) at ($(nn.east)+(0.4,0)$) {\(\hat{y}^{(i)}\): fish};
    \node (a1) at (0,-0.65) {By asking how much the network will change \(\hat{y}^{(i)}\) if};
    \node (b1) at ($(a1.west)+(0,-\h)$) {removing a certain training image, we can find:};
    \node (c1) at ($(b1.west)+(0,-0.65)$) {most helpful\textsuperscript{2} training images:};
    \node (d1) at ($(c1.east)+(0.08,0)$) {\includegraphics[width=0.8cm]{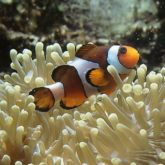}};
    \node (e1) at ($(d1.east)+(0.15,0)$) {\includegraphics[width=0.8cm]{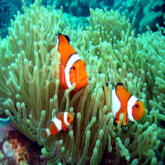}};
    \node (f1) at ($(e1.east)+(0.05,0)$) {\cite{koh2017ubbpv}};
\end{tikzpicture}